%% file: paper3.tex
\documentclass[pdftex]{article}
\usepackage[dvipdfmx]{graphicx} 
\usepackage{tikz}
\usepackage{amssymb}
\usepackage{color}
\usepackage{plain}
\newtheorem{teiri}{Theorem}
\newtheorem{prop}{Proposition}

\newtheorem{algo}{Algorithm}

\def\ci{\perp\!\!\!\perp}
\def\ci{\perp\!\!\!\perp}
\title{Causal Order Identification to Address Confounding:\\ Binary Variables}
\author{Joe Suzuki\thanks{prof.joe.suzuki@gmail.com} and Yusuke Inaoka}
\date{}
\begin{document}
\maketitle
\begin{abstract}
This paper considers an extension of the linear non-Gaussian acyclic model (LiNGAM) that determines the causal order among variables from a dataset when the variables are expressed by a set of linear equations, including noise. In particular, we assume that the variables are binary.
The existing LiNGAM assumes that no confounding is present, which is restrictive in practice.
Based on the concept of independent component analysis (ICA), this paper proposes an extended framework in which the mutual information among the noises is minimized.
Another significant contribution is to reduce the realization to the shortest path problem, in which the distance between each pair of nodes expresses an associated mutual information value, and the path with the minimum sum (KL divergence) is sought.
Although $p!$ mutual information values should be compared, this paper dramatically reduces the computation when no confounding is present.
The proposed algorithm finds the globally optimal solution, while the
existing approaches locally greedily seek the order based on
hypothesis testing. We use the best estimator in the sense of Bayes/MDL that correctly detects independence for mutual information estimation.
Experiments using artificial and actual data show that the proposed version of LiNGAM achieves significantly better performance, particularly when confounding is present.
\end{abstract}

\section{Introduction}

Suppose that we have data $x^n$ and $y^n$ of size $n$ for variables $X$ and $Y$, respectively.
If we write $X\rightarrow Y$ to denote that $X$ and $Y$ are the cause and effect, respectively,
one might think to choose one of the $X\rightarrow Y$ and $Y\rightarrow X$
by comparing the likelihood.
However, this reasoning fails to identify the direction of causality because the likelihoods are identical (Markov equivalence).
We may construct a skeleton from structure learning procedures such as the PC algorithm \cite{pc} and greedy equivalence search \cite{ges}.
However, in general, we cannot finalize the causal order among the variables.
In this paper, we are interested in seeking the causal order rather than the skeleton.

In 2006, Shimizu \cite{shimizu06} proposed a novel criterion:
if $X\rightarrow Y$,
there should be a function $f$ and a variable $\epsilon$ such that $Y=f(X,\epsilon)$ and $X$ and $\epsilon$ are independent.
For example, we can determine the causal direction by identifying which of
$X\ci \epsilon$ and $Y\ci \epsilon'$ is correct in
\begin{equation}\label{eq81}
Y=aX+\epsilon
\end{equation}
\begin{equation}\label{eq82}
X=a'Y+\epsilon'
\end{equation}
for some $a,a'\in {\mathbb R}$.
However, both independences may hold for some $a,a'$, and we cannot
distinguish between $X\rightarrow Y$ and $Y\rightarrow X$ in this case.
Shimizu \cite{shimizu06} proved that such an inconvenience does not occur if and only if either $X$ or $\epsilon$ is non-Gaussian
when the true model is (\ref{eq81}).

The theory that we refer to as {\it LiNGAM} (linear non-Gaussian acyclic model \cite{shimizu11,apo13}) in this paper was inspired by independent component analysis (ICA). ICA finds independent component variables such as $X$ and $\epsilon$
from data $x^n$ and $y^n$ of $X$ and $Y$.
There are two versions of LiNGAM---ICA-LiNGAM \cite{shimizu06} and direct-LiNGAM \cite{shimizu11}---and both are based on ICA.

The main problem with LiNGAM is that it assumes that no confounding is present.
The reasoning in LiNGAM follows because it assumes that exactly one of\footnote{We write $X\ci Y$ when $X$ and $Y$ are independent.}
$X\ci \epsilon$ in (\ref{eq81}) and $Y\ci \epsilon'$ in (\ref{eq82})
is true. However, this constraint makes LiNGAM restrictive in practice.
In fact, as the number of variables increases, the assumption will not be satisfied.
For example, if $X=\epsilon_1$, $Y=aX+\epsilon_2$, and $Z=bX+cZ+\epsilon_3$, with $a,b,c\in {\mathbb R}$, are true,
then, we must require the noises $\epsilon_1,\epsilon_2,\epsilon_3$ to be independent, which is rather unrealistic.

In this paper, we propose an extension of LiNGAM that can address the case in which confounding is present among discrete variables \cite{ina11, peters11}.
In particular, we relax the constraint in LiNGAM and only assume that the true structure minimizes the noises' mutual information.
The idea of minimizing the mutual information among the independent components was proposed in ICA, so the extension seems reasonable.
Some authors have proposed ways to avoid the effects of confounders without extending LiNGAM.
However, these methods require the knowledge that confounding is present a priori and take an exponential time of the number $p$ of variables \cite{chen13, tashiro14}.
Besides, Shimizu et al. \cite{shimizu14} considered individual-specific effects that are sometimes the
source of confounding, and proposed an empirical Bayesian approach for estimating possible causal direction. 

One contribution of this paper is to propose an efficient procedure to achieve this goal.
Although LiNGAM searches the variable order using hypothesis testing in a greedy manner, we search for the globally optimal order based on the shortest path problem, assuming direct-LiNGAM \cite{shimizu11}.
For the three-variable case, we minimize the (total) mutual information
$$I(e_1,e_2,e_3)=I(e_1,\{e_2,e_3\})+I(e_2,e_3).$$
There are six paths for $p=3$, and each has mutual information; the corresponding mutual information value is assigned as a distance for each pair of connected nodes.
We choose the path with the minimum mutual information.
In particular, we prove that the computation almost surely completes as fast as the original LiNGAM if no confounding is present.

Another issue is the estimation of mutual information.
In this paper, we assume that the variables are binary.
Then, one might think that the estimator can be constructed as a function of the relative frequencies.
However, the maximum likelihood estimator tends to show larger estimates than the true mutual information value due to overfitting.
We apply an optimal estimator based on the Bayes/MDL criteria \cite{uai93}.

We admit that the binary LiNGAM \cite{ina11} and \cite{peters11} is rarer than the continuous one. 
However, the binary LiNGAM can be used in any binary dataset, such as the Asia dataset by S. Lauritzen \cite{asia}. 
Our goal in the future is to propose LiNGAM for categorical data rather than binary data,
and we regard the current work as its first step. 

In general, it is possible that two variables cannot be not ordered. For example, suppose that both of 
$X_1\rightarrow X_2 \rightarrow X_3 \rightarrow X_4$
and $X_1\rightarrow X_3 \rightarrow X_2 \rightarrow X_4$ are consistent with the true order when 
$\{X_2,X_3\}$ is after $X_1$ and before $X_4$ and  
the order between $X_2$ and $X_3$ does not matter.
The proposed algorithm outputs one of the possible orders based on the data.

Our contributions include the following:
\begin{enumerate}
\item formulate LiNGAM based on minimizing the mutual information value and make LiNGAM available even when confounding is present;
\item reduce finding the optimal causal order to the shortest path problem in the ordered graph such that the distances are the mutual information values between the nodes;
\item apply mutual information estimation based on the Bayes/MDL criteria; and
\item find that the proposed LiNGAM achieves significantly better performance than the original LiNGAM for any case (with and without confounding).
\end{enumerate}

This paper is organized as follows:
Section 2 explains the background for understanding the results and discusses existing works.
Section 3 states the results of this paper, in particular the principle and procedure.
Section 4 shows an example and experiments and examines the
effectiveness of our approach.
Section 5 concludes this paper with a summary of the results and future work.

\section{Preliminaries}
In this section, we provide essential background knowledge for understanding the results in later sections.

\subsection{LiNGAM for two continuous variables}

Given actually occurring sequences $x^n:=(x_1,\ldots, x_n), y^n:=(y_1,\ldots, y_n)\in {\mathbb R}^n$ of length $n \geq 1$,
we wish to estimate which of the variables $X$ and $Y$ are the cause and effect.
If we write $X\rightarrow Y$ to denote that $X$ and $Y$ are the cause and effect, respectively,
LiNGAM is a criterion for determining the direction of ``$\rightarrow$''.

For simplicity, we assume that the expectations of $X$ and $Y$ are both zero.
If one of them is expressed by a linear regression of the other, then the problem reduces to identifying either of the two regressions:
\begin{equation}\label{eq32}
\left\{
\begin{array}{l}
X=e_1\\
Y=aX+e_2
\end{array}
\right.
\end{equation}
or
$\displaystyle \left\{
\begin{array}{l}
Y=e_1'\\
X=a'Y+e_2'
\end{array}
\right.$
for some $a,a'\in {\mathbb R}$ such that the variables
$\{e_1,e_2\}$ and $\{e_1',e_2'\}$ are independent, and the expectations of the four variables are zero.
If the true regression is expressed by one of the two, the problem further reduces to which of $e_1 \ci e_2$ and $e_1' \ci e_2'$ is more likely.

However, we may not be able to distinguish between $e_1\ci e_2$ and $e_1' \ci e_2'$.
We know that for Gaussian variables, independence and zero correlation are equivalent.
It is known \cite{skito,darmois} that given $a\in {\mathbb R}$ and $e_1\ci e_2$,
there exists $a'\in {\mathbb R}$ such that $e_1'\ci e_2'$ if and only if both $e_1$ and $e_2$ are Gaussian.
Hereafter, LiNGAM assumes that at least one of $e_1$ and $e_2$ does not follow a Gaussian distribution.
\begin{prop}[Shimizu et al. \cite{shimizu11}]\rm \label{prop1}
Suppose that $X$ and $Y$ are not independent.
The following two conditions are equivalent:
\begin{enumerate}
\item both $X$ and $Y$ are Gaussian.
\item there exist $a,a'\in {\mathbb R}$ such that both $X\ci (Y-aX)$ and $Y\ci (X-a'Y)$.
\end{enumerate}
\end{prop}

\subsection{LiNGAM for multiple continuous variables}

Given occurring sequences $x^n:=(x_1,\ldots, x_n), y^n:=(y_1,\ldots, y_n), z^n:=(z_1,\ldots, z_n)\in {\mathbb R}^n$ of length $n$, we wish to estimate the order of the variables $X, Y, Z$ such that one is the cause of the other two, and of these two, one is the cause of the other.
LiNGAM even determines the direction ``$\rightarrow$'' for more than two variables.

We assume that the expectations of $X$, $Y$, and $Z$ are zero for simplicity.
There are $3!=6$ orders for $X,Y,Z$.
For example, if $X\rightarrow Y\rightarrow Z$ is true, then we assume that they are generated by
\begin{equation}\label{eq12}
\left\{
\begin{array}{l}
X=e_1\\
Y=aX+e_2\\
Z=bX+cY+e_3
\end{array}
\right.
\end{equation}
for some $a,b,c\in {\mathbb R}$ such that the variables $e_1, e_2, e_3$ have expectation zero and are independent.

We first compute the six quantities 
$$x_y^n:=x^n-\frac{c(y^n,x^n)}{v(y^n)}y^n\ ,\ y_x^n:=y^n-\frac{c(x^n,y^n)}{v(x^n)}x^n$$
$$z_x^n:=z^n-\frac{c(x^n,z^n)}{v(x^n)}x^n\ ,\ z_y^n:=z^n-\frac{c(y^n,z^n)}{v(y^n)}y^n\ ,$$
$$x_z^n:=x^n-\frac{c(z^n,x^n)}{v(z^n)}z^n\ ,\
{\rm and}\ y_z^n:=y^n-\frac{c(z^n,y^n)}{v(z^n)}z^n\ ,$$
and compare the independence of
$\{x^n, (y_x^n,z^n_x)\}$, $\{y^n, (z_y^n,x^n_y)\}$, and $\{z^n, (x_z^n,y_z^n)\}$.

If $X$ is chosen as the cause in the first stage,
we compare the two pairs $\{y^n_x,z^n_{xy}\}$ and $\{z^n_x,y_{zx}^n\}$
and choose the pair that is more independent,
where $$z^n_{xy}:= z_x^n-\frac{c(y_x^n,z_x^n)}{v(y_x^n)}y_x^n\ ,\ y^n_{xz}:= y_x^n-\frac{c(z_x^n,y_x^n)}{v(z_x^n)}z_x^n\ .$$

When we have $n$ samples for $p$ variables, we can similarly determine the order.
If one source is Gaussian among the noises $e_1,\ldots,e_p$, the reasoning above follows.

\subsection{Confounding}

We say that confounding exists if the noises $e_1,e_2,\ldots,e_p$ are not independent.
More precisely, confounding exists if for the noises $e_1,e_2,\ldots,e_p$,
$$P_{1,\ldots,p}(e_1=\epsilon_1, \cdots, e_p=\epsilon_p)\not=P_1(e_1=\epsilon_1)\cdots P_p(e_p=\epsilon_p)$$
with non-zero probability w.r.t. noise values $\epsilon_1,\ldots,\epsilon_p\in {\mathbb R}$ for any order of the $p$ variables $e_1,e_2,\ldots,e_p$.
Although the definition might be different from the other literature, we consider such cases in this paper.
For example, it assumes that
$\{e_1,e_2\}$ and $\{e_1,e_2,e_3\}$ are true in (\ref{eq32}) and (\ref{eq12}), respectively,
if they are the true models.

\subsection{ICA}

In general, the mutual information $I(U, V)$ between $U$ and $V$ is
often used to measure how mutually dependent $U$ and $V$ are.
For example, for the noises $e_1,e_2$, we have $I(e_1,e_2)=0 \Longleftrightarrow e_1\ci e_2$.

By {independent component analysis}, ICA \cite{ica,comon94}, we mean
to minimize the mutual information $I(s_1,s_2)$ between
variables $s_1,s_2$ such that 

$\displaystyle
\left[
\begin{array}{c}
X\\
Y
\end{array}
\right]
=A
\left[
\begin{array}{c}
s_1\\
s_2
\end{array}
\right]
$ for a matrix $A=
\left[
\begin{array}{cc}
a_{11}&a_{12}\\
a_{21}&a_{22}
\end{array}
\right]
$, given variables $X,Y$, where the row and column vectors are nonzero.
One can check that the problem reduces to finding $A$ in the form
$\displaystyle
A=
\left[
\begin{array}{cc}
1&a'\\
a&1
\end{array}
\right]
$
for some $a,a'$.
On the other hand, LiNGAM chooses either
$\displaystyle
A=\left[
\begin{array}{cc}
1&0\\
a&1
\end{array}
\right]
$ or $
\displaystyle
A=\left[
\begin{array}{cc}
1&a'\\
0&1
\end{array}
\right]
$
for some $a,a'\not=0$.
In this sense, LiNGAM solves a restricted case of ICA, as the authors of LiNGAM remarked in their initial version \cite{shimizu06}.

\subsection{LiNGAM for binary variables}

In this paper, we consider LiNGAM for discrete variables.
Binary LiNGAM was independently considered by Peters et al. \cite{peters11} and Inazumi et al. \cite{ina11}.

Suppose that $X,Y$ take binary values (in $\{0,1\}$) and that they are stochastically related by
\begin{equation}\label{eq42}
\left\{
\begin{array}{l}
X=e_1\\
Y=f(X)+e_2
\end{array}
\right.
\end{equation}
with $f: \{0,1\}\rightarrow \{0,1\}$, where $e_1,e_2$ randomly takes binary values,
and "$+$" denotes the exclusive-or operation.
Note that $f(\cdot)$ is either of $0$,1,$\cdot$,$\cdot+1$. If $f(X)$ is either 0 or 1, then $X\ci Y$, which means
that both $e_1\ci e_2$ and $e_1'\ci e_2'$ occur.
The problem is to identify
the order (whether (\ref{eq42}) or
$\displaystyle \left\{
\begin{array}{l}
Y=e_1'\\
X=f'(Y)+e_2'
\end{array}
\right.$
) and the function $f,f': \{0,1\}\rightarrow \{0,1\}$
for binary random variables $e_1,e_2,e_1',e_2'$, given data $x^n, y^n \in \{0,1\}^n$.
They found the order and function by identifying which is more likely between $e_1\ci e_2$ and $e_1'\ci e_2'$,
assuming that there exist no confounders, i.e., either $e_1\ci e_2$ or $e_1'\ci e_2'$ is true.

The arithmetic is excluseive-or and all the variables and coefficients are either zero or one. Note that $2x=0$, $x^2=x$, and $x+y=x-y$, 
for $x,y=0,1$ (all the arithmetic is modulo two). The residues are computed as $y.x=y-x$ and $y.x=y$ for $Y=X+e$ and $Y=e$, respectively. 
The coefficient $a(y,x)=0,1$ such that $y.x=y-a(y,x)x$ is determined by which of $(x,y-x)$ and $(x,y)$ is closer to independence.
In general,
if we define $f(u,v):=u-a(u,v)v$ and $x.yz:=f(x.y,z.y)$, then we recursively obtain
$x.y=f(x,y)=x-a(x,y)y$,
$z.y=f(z,y)=z-a(z,y)y$, and 
\begin{eqnarray*}
x.yz
&=&f(x.y,z.y)=x.y-a(x.y,z.y)z.y=\{x-a(x,y)y\}-a(x.y,z.y)\{z-a(z,y)y\}\\
&=&\{x-a(x,y)y\}-a(x-a(x,y)y,z-a(z,y)y)\{z-a(z,y)y\}
\end{eqnarray*}

They proved a similar identifiability as Propotion \ref{prop1}:
\begin{prop}[Inazumi et al. \cite{ina11}]\rm \label{prop2}
Suppose that $X$ and $Y$ are not independent.
If both $X$ and $Y$ take zeros and ones equiprobably, 
then there exist $f,f': \{0,1\}\rightarrow \{0,1\}$ such that both $X\ci (Y-f(X))$ and $Y\ci (X-f'(Y))$.
\end{prop}

We are not concerned about whether $f(X)=X$ or $f(X)=X+1$ in deciding whether $X\rightarrow Y$ or $Y\rightarrow X$
because $e_1 \ci e_2 \Longleftrightarrow (e_1+1)\ci (e_2+1)$, where "$+$" expresses the exclusive-or operation.
Thus, as long as $X$ and $Y$ are not independent,
we may assume that $f(X)=X$.

For continuous variables, the HSIC (Hilbert Schmidt independence criterion) \cite{hsic} is often used to test the independence of two variables.
For the discrete variables, the authors \cite{ina11,peters11} used the G-test or its variant based on mutual information.
Note that the previous LiNGAM approach assumes that no confounding exists, which is unrealistic in reality, and that the order identification process is done greedily.

\section{An Extended Criterion for Addressing Confounding}

\subsection{Minimizing the Mutual Information for Identifying the
  Order}

We have seen that LiNGAM prevents us from addressing confounding and imposes a restrictive condition:
the noises $e_1,\ldots,e_p$ among the $p$ variables should be independent.
For this reason, we may say that LiNGAM does not provide any reliable results for actual data.

We propose a relaxed criterion based on ICA: the true model should minimize the mutual information $I(e_1,\ldots,e_p)$ defined by
\begin{equation}\label{eq7-7}
\sum_{\epsilon_1}\cdots \sum_{\epsilon_p} P_{1,\ldots,p}(\epsilon_1,\ldots,\epsilon_p)
\log \frac{P_{1,\ldots,p}(\epsilon_1,\ldots,\epsilon_p)}{P_{1}(\epsilon_1)\cdots P_{p}(\epsilon_p)}\ ,
\end{equation}
where $P_{1,\ldots,p}$ and $P_i$, $i=1,\ldots,p$, are the associated probabilities of the noises
$e_1,\ldots,e_p$ and the values $\epsilon_1,\ldots,\epsilon_p$ range over the values that the noises take.
For example, for the noises $e_1,e_2,e_3$ in the previous section, we have
$$I(e_1,e_2,e_3)=0 \Longleftrightarrow e_1\ci \{e_2,e_3\}, e_2\ci e_3$$
because $I(e_1,e_2,e_3)=I(e_1,\{e_2,e_3\})+I(e_2,e_3)$.

Note that $e_1,\ldots,e_p$ being independent implies $I(e_1,\ldots,e_p)=0$, although the converse does not hold.
In this sense, the criterion is consistent with LiNGAM: if the variables satisfy the assumption of LiNGAM and the order is optimal,
it is also optimal for the novel criterion. However, even if the assumption does not hold,
order identification works under the novel criterion.

We say that $X_1,\ldots,X_p$ follows an additive noise model \cite{Kano}
\begin{equation}\label{eq7-18}
X_{i}=f_i(X_1,\ldots,X_{i-1})+e_i
\end{equation}
with the function $f_i$ being known, $i=1,\ldots,p$, such that $e_1,\ldots,e_p$ are independent.
Suppose that we are given $p$ variables whose noises $e_1,\ldots.e_p$ may not be independent.
If we wish to know which among the $p!$ additive noise models fits,
we change the order of the given $p$ variables
and evaluate the Kullback-Leibler (KL) divergence (\ref{eq7-7}) for
$P(e_i=1)=\sum P(e_1,\ldots,e_{i-1},1,e_{i+1},\ldots,e_p)$ and  $P(e_i=0)=1-P(e_i=1)$, where
the sum ranges over $(e_1,\ldots,e_{i-1},e_{i+1},\ldots,e_p)\in \{0,1\}^{p-1}$, $i=1,\ldots,p$.
We choose the additive noise model such that the KL divergence is minimized
and evaluate how significant the confounding is using the KL divergence value.

The criterion is easy to accept because the idea is based on the ICA from which LiNGAM was derived.
In this section, we mention several merits of applying the novel criterion for discrete variables.

\subsection{Finding the Optimal Variable Order}

In this paper, we define the following quantity with respect to $X,Y,Z$ given $x^n,y^n,z^n$:
\begin{equation}\label{eq14}
I_n(x^n,y_x^n,z_{xy}^n):=I_n(x^n,\{y_x^n,z_{xy}^n\})+I_n(y_x^n,z_{xy}^n),
\end{equation}
where $I_n()$ is an estimate of $I()$ given $n$ samples.
We have observed that $e_1^n:=x^n$, $e_2^n=y_x^n$, and $e_3^n:=z_{xy}^n$ are generated by $e_1,e_2,e_3$ in (\ref{eq12}); therefore, we regard (\ref{eq14}) as the estimate of $I(e_1,e_2,e_3)$.
Moreover, we compare (\ref{eq14}) with the other five quantities
to determine which order is more likely. If the noise set $\{e_1,e_2,e_3\}$ is independent,
then $I(e_1,e_2,e_3)=0$, and $I_n(x^n,y_x^n,z_{xy}^n)$ converges to zero as $n$ grows.
On the other hand, if the noise set $\{e_1',e_2',e_3'\}$ for $Y\rightarrow Z\rightarrow X$ is not independent,
then $I(e_1',e_2',e_3')>0$, and $I_n(y^n,z_y^n,x_{yz}^n)$ converges to a positive value.

\begin{figure*}
\input{figure007}
\caption{\label{fig03}
The ordered graph consists of the subsets of $V$, where the blue rectangles and red paths are the opened nodes and solutions.
}
\end{figure*}
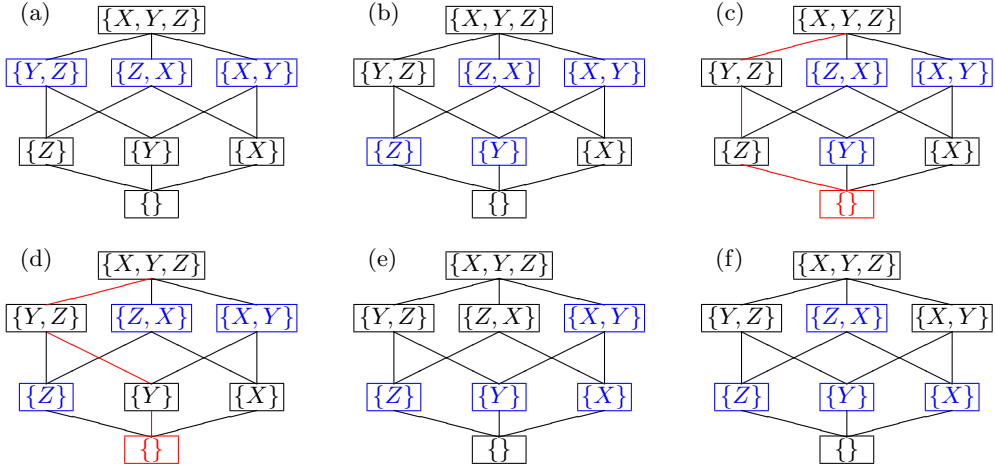

In this paper, we propose a procedure to find the shortest path for the problem.
For ease of understanding, we consider the simplest case ($p=3$) with eight nodes,
$$\{X,Y,Z\}, \{Y,Z\}, \{Z,X\}, \{X,Y\},
\{X\}, \{Y\}, \{Z\}, \{\},$$
and twelve edges, as in Figure \ref{fig03}.
Suppose that we have DATA=$\{x^n,y^n,z^n\}$ as input.
Then, we can compute the residues and
a mutual information estimate value for each of the twelve edges
(we assume that $I_n(\{x^n_{yz}\},\{\})=I_n(\{y^n_{zx}\},\{\})=I_n(\{z^n_{xy}\},\{\})=0$).

We regard the mutual information estimation values as the distances.
Then, for each node $v$, we can compute
the length $d(v)$ of the path from the top $\{X, Y, Z\}$ to $v$ and the sum of the distances of the edges along the path.
If more than one path exists to a node, we choose the shortest path and store it in the node.
For example, for the path $\{X,Y,Z\}\rightarrow \{Y,Z\}\rightarrow \{Z\}\rightarrow \{\}$,
the sum of the distances is
\begin{eqnarray*}
&&I_n(x^n,\{y_x^n,z_x^n\})+I_n(y_x^n,z_{xy}^n)+0\\
&=&I_n(x^n,\{y_x^n,z_{xy}^n\})+I_n(y_x^n,z_{xy}^n)+0\\
&=&I(x^n,y_x^n,z_{xy}^n)\ ,
\end{eqnarray*}
which is the estimated mutual information value of the noises $e_1,e_2,e_3$ such that $X=e_1,Y=aX+e_2,Z=bX+cY+e_3$ for some constants $a,b,c$.
Our goal is to find the shortest path from the top $\{X, Y, Z\}$ to the bottom $\{\}$.

First, we compute the lengths of the edges from the top $\{X,Y,Z\}$ to $\{Y,Z\},\{Z,X\},\{X,Y\}$:
$$d(\{Y,Z\}):=I_n(x^n,\{y_x^n,z_x^n\})\ ,$$
$$d(\{Z,X\}):=I_n(y^n,\{z_y^n,x_y^n\})\ ,\ {\rm and}$$
$$d(\{X,Y\}):=I_n(z^n,\{x_z^n,y_z^n\})\ .$$
We close the top node $\{X,Y,Z\}$ and open $\{Y,Z\}, \{Z,X\}, \{X,Y\}$ (Figure \ref{fig03} (a)).
Suppose that $d(\{Y, Z\})$ is the smallest of the three nodes.
Then, we compute $I_n(y_x^n,z_{xy}^n)$ and $I_n(z_{x}^n,y_{zx}^n)$ and obtain
\begin{equation}\label{eq17}
d(\{Z\}):=d(\{Y,Z\})+I_n(y_x^n,z_{xy}^n)
\end{equation}
and
$d(\{Y\}):=d(\{Y,Z\})+I_n(z_{x}^n,y_{zx}^n)$,
respectively. We close $\{Y,Z\}$ and open $\{Z\}$ and $\{Y\}$
(Figure \ref{fig03} (b)).

If $d(\{Z\}\})$ is the smallest in Figure \ref{fig03} (b), then $X\rightarrow Y\rightarrow Z$ is the shortest path
(Figure \ref{fig03} (c)); if $d(\{Y\})$ is the smallest in Figure \ref{fig03} (b), then $X\rightarrow Z\rightarrow Y$ is the shortest path (Figure \ref{fig03} (d)).
On the other hand, if $d(\{Z,X\})$ is the smallest in Figure \ref{fig03} (b),
we compute $I_n(z_y^n,x_{yz}^n)$ and $I_n(x_{y}^n,z_{xy}^n)$, and we obtain
$d(\{X\}):=d(\{Z,X\})+I_n(z_y^n,x_{yz}^n)$
and
\begin{equation}\label{eq18}
d(\{Z\}):=d(\{Z,X\})+I_n(x_{y}^n,z_{xy}^n)\ .
\end{equation}
We close $\{Z,X\}$ and open $\{X\}$ and $\{Z\}$.
However, the values of (\ref{eq17}) and (\ref{eq18}) conflict; thus, we replace (\ref{eq17}) with (\ref{eq18}) if (\ref{eq18}) is smaller (Figure \ref{fig03} (e)).
Finally, if $d(\{X,Y\})$ is the smallest in Figure \ref{fig03} (b),
we obtain the state depicted in Figure \ref{fig03} (f), in which the values of $d(\{Y\})$ conflict, and the shorter path is chosen from $\{X,Y,Z\}$ to $\{Y\}$.

We continue this procedure to obtain the distance $d(\{\})$ and the shortest path
from the top $\{X,Y,Z\}$ to the bottom $\{\}$.

We have the following procedure (Algorithm \ref{algo1}) with input DATA and output SHORTEST\_PATH.
Let TOP and BOTTOM be the top and bottom nodes, and we define $append((u_1,\ldots,u_s), u_{s+1}):=(u_1,\ldots,u_s,u_{s+1})$
for the nodes $u_1,u_2,\ldots,u_{s+1}$.
\begin{algo}\rm \label{algo1}
Let $OPEN:=\{TOP\}$, CLOSE:=$\{\}$, ${\rm path}({\rm TOP}):=()$, $r({\rm TOP}):=$DATA, and repeat the following:
\begin{enumerate}
\item Move node $v\in$ OPEN to CLOSE such that $d(v)$ is the smallest among the nodes in OPEN,
and suppose that the nodes $v_1,\ldots,v_m$ are connected to $v$;
\item If BOTTOM $\in$ OPEN, SHORTEST\_PATH$=append({\rm path}(v),\{\})$ and terminate;
\item For each $i=1,\ldots,m$:
\begin{enumerate}
\item If $v_i \not\in$ OPEN, compute the residue $r(v_i)$ of $v_i$ from $r(v)$;
\item Compute the mutual information estimation $mi$ via $r(v)$ and $r(v_i)$.
\item If either $v_i \not\in$ OPEN or \{$v_i \in$ OPEN, and $d(v)+mi<d(v_i)$\}, then $d(v_i)=d(v)+mi$ and
${\rm path}(v_i)=append({\rm path}(v),v_i)$.
\item join $v_i$ to  OPEN if $v_i\not\in OPEN$ for $j=1,\ldots,m.$
\end{enumerate}
\end{enumerate}
\end{algo}
Note that Algorithm 1 does not compute the residues and mutual information estimations initially; instead, it calculates each step by step when necessary to reduce computational complexity.
In addition, SHORTEST\_PATH is expressed by a sequence of nodes such as $(\{X,Y,Z\},\{Y,Z\},\{Z\},\{\})$ rather than variables separated by arrows, as in $X\rightarrow Y\rightarrow Z$.

\begin{teiri}\rm
Algorithm 1 computes the order of the variables that minimize an estimate of the KL divergence defined by (\ref{eq7-7}).
\end{teiri}

\begin{figure*}
\begin{center}
\begin{tabular}{rl}
\input{nc_rate1}&\input{nc_rate2}\\
\end{tabular}
\end{center}
\caption{\label{fig002} For any combination of $(n,p)$, in the complete matching criterion,
the proposed method significantly outperforms the conventional method that seeks the order greedily.
However, for the pairwise criterion, the performances are not as significant, which is due to the nature of the global and local searches.}
\end{figure*}
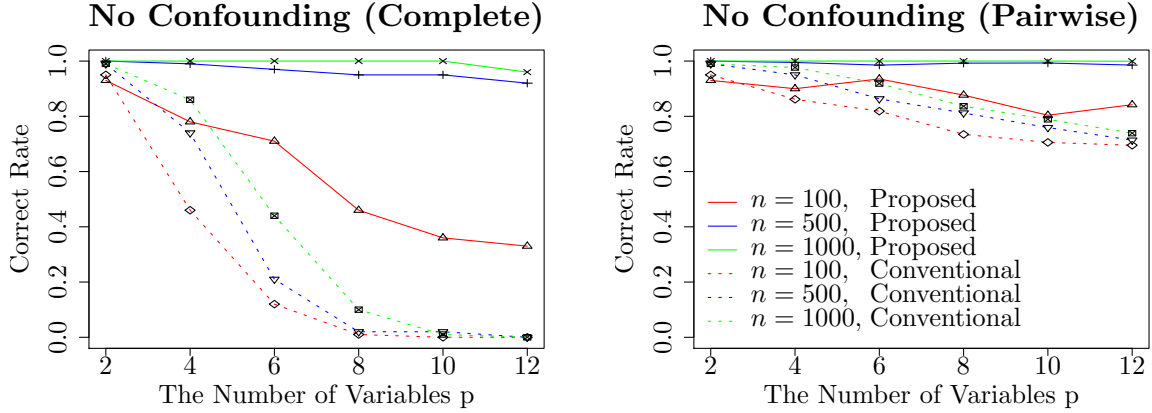

\subsection{Independence Testing and Mutual Information}

In LiNGAM for binary variables, independence testing based on hypothesis testing may be the best choice.
However, if confounding exists, hypothesis testing does not work for our purpose because our goal is to order the $p$ variables that minimize (\ref{eq14}).
Even when no confounding exists, we anticipate that the existing
search performs worse because the decision is made in a greedy
manner. If we globally minimize (\ref{eq14}) rather than locally
minimizing test statistics such as $G$-statistics at each stage,
we can postpone the decision of the whole order to the end of the search:
even if we make an error in the earlier stages,
if it detects that the intermediate result is not consistent with the decisions in the later stage,
the shortest path search finds the local error and can make a globally correct decision.

For estimating mutual information $I(U,V)$ values given data $u^n\in L^n$ and $v^n \in M^n$,
the most common approach is
\begin{equation}\label{eq71}
I_n:=\sum_{u\in L}\sum_{v\in M} \frac{c_{UV}(u,v)}{n}\log \frac{c_{UV}(u,v)/n}{c_U(u)/n\cdot c_V(v)/n}\ ,
\end{equation}
where $c_{UV}(\cdot,\cdot), c_{U}(\cdot), c_{V}(\cdot)$ are the associated counters.
However, (\ref{eq71}) overestimates the true value $I(U,V)$. In fact, the quantity
$$n\{I(U,V)-I_n(u^n,v^n)\}-\frac{(\alpha-1)(\beta-1)}{2}\log n$$
is almost surely bounded. Hence, we may use a consistent estimator \cite{uai93}
\begin{equation}\label{eq72}
J_n:=\max\{0,I_n-\frac{(\alpha-1)(\beta-1)}{2n}\log n\}
\end{equation}
that almost surely satisfies $J_n=0 \Longleftrightarrow U\ci V$ as $n\rightarrow \infty$.
For one and $1\leq q\leq p-1$ variables, we have $\alpha=2^1$ and $\beta=2^q$.

The estimate $I_n$ in (11) overestimates the mutual information and is positive with nonzero probability even when
$U,V$ are independent.  
Although both $I_n$ and $J_n$ converges to zero,
$I_n$ keeps positive while $J_n$ becomes exactly zero except finite $n$.
In this sense, $J_n$ can detect whether the true mutual information is zero or not while $I_n$ cannot.

Although the second term in (\ref{eq71}) almost surely converges to the true value,
the value is always a (small negligible) positive due to overfitting.
When no confounder exists, if we use (\ref{eq72}) instead, then the values along the true path are almost surely always chosen
because those values are almost surely estimated as zero.
Thus, the total number of opened nodes (the number of mutual information computations) is $p+(p-1)+\cdots+1=p(p-1)/2$.
On the other hand, if (\ref{eq71}) is used, even if the true value is estimated to be positive, we may require more computation.

\begin{figure*}
\begin{center}
\begin{tabular}{rl}
\input{ec_rate1}&\input{ec_rate2}\\
\end{tabular}
\end{center}
\caption{\label{fig5} Even when confounding exists, the proposed
  method still performs better, particularly for the complete matching criterion.}
\end{figure*}
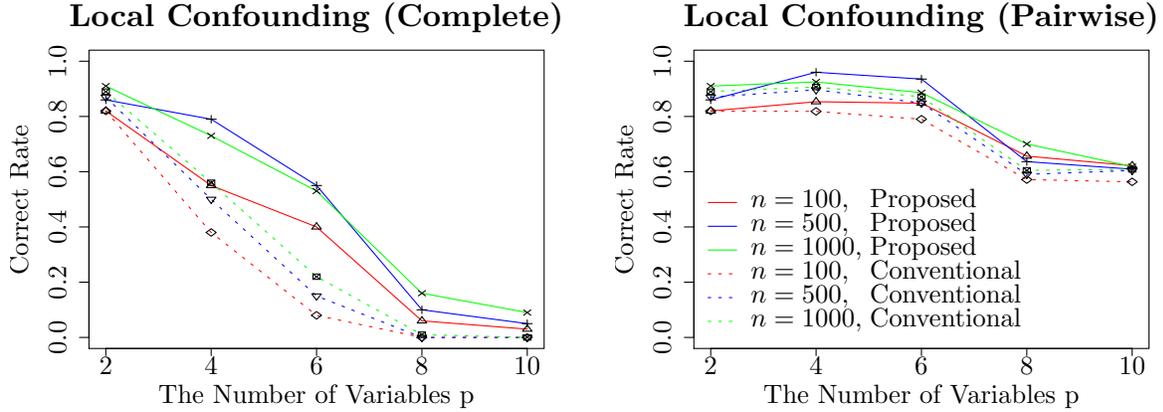

\begin{teiri}\rm
If no confounder exists, at most $p(p-1)/2$ mutual information values out of $p(2^{p-1}-1)$ are computed with probability one as $n\rightarrow \infty$.
\end{teiri}
(The worst-case requires exponential order computation of $p$, which seems to be very rare.)

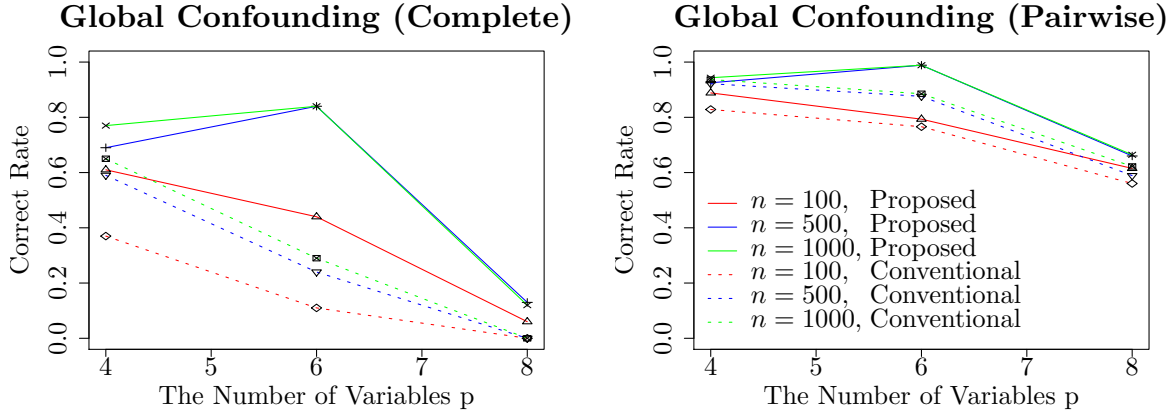
\begin{figure*}
\begin{center}
\begin{tabular}{rl}
\input{ec2_rate1}&\input{ec2_rate2}\\
\end{tabular}
\end{center}
\caption{\label{fig4} The proposed method performs better not only in
  the no- and local-confounding cases but also in the global-confounding case.}
\end{figure*}

\begin{table*}
\caption{How many times the mutual information values were estimated during the order identification process on average.
From the table, we find that even if confounding exists, the number of
mutual information computations was not substantially increased.\label{tab3}
}
\begin{center}
\input{tab3}
\end{center}
\end{table*}

\begin{figure*}
\input{figure008}
\caption{
The symbols $\bigcirc$ and $\times$ in the three tables denote whether each pair of variables can be ordered by
LvLiNGAM \cite{entner13}, ParceLiNGAM \cite{tashiro14}, and the extended (proposed) LiNGAM for the left figure.
LvLiNGAM \cite{entner13} 
obtains the orders of variable pairs that are not affected by any confounder,
and estimate the order of the whole variables by combining those pairwise orders.
ParceLiNGAM \cite{tashiro14} 
divides the variable set into the upper, middle, and lower variable sets 
by top-down and bottom-up causal searches, where the upper and lower variable sets are 
the maximal subsets that contain no confounder but the top and bottom variables, respectively. 
For LvLiNGAM, because $X$ is not affected by any confounder, it detects 
$X\rightarrow Y$,
$X\rightarrow Z$,
$X\rightarrow T$,
$X\rightarrow W$. However, each of $\{Y, T\}$ and $\{Y, W\}$ is affected by $f$, and each of $\{Z, W\}$ and $\{Z, T\}$ is affected by $g$ so that LvLiNGAM can not order the four pairs. 
For ParceLiNGAM, because it considers subsets such as $\{Y,Z,T\}$, it detects additional orders $Y\rightarrow T$ and $Z\rightarrow T$.
However, the proposed procedure can order all the pairs, including the
latter two. \label{tab02}}
\end{figure*}
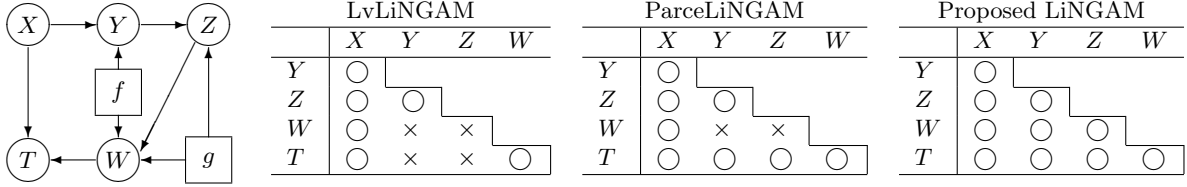

ParceLiNGAM \cite{tashiro14} and LvLiNGAM \cite{entner13} are major existing approaches to LiNGAM when confounding is present.
For the details on them, see the references.
We cannot simply compare the proposed (extended) LiNGAM to the two existing LiNGAM approaches for addressing confounding
because the former and latter proceed with discrete and continuous
variables. 

From Figure \ref{tab02}, LvLiNGAM and ParceLiNGAM work under limited conditions and require us to know a priori that confounding exists.
Moreover, if there is no confounding, they require much more computation than the existing LiNGAM.
The proposed method can be used for cases with and without confounding, which is of significant merit.

\section{Experiments}

We examined the performance of the proposed method via artificial and actual data.
To analyze its performance in detail, we define two criteria for the correctness of the order identifications:
the complete and pairwise matching rates.
Suppose that the estimated and true orders are $a_1,\ldots,a_p$ and $b_1,\ldots,b_p$ for the $p$ variables.
The complete matching evaluates whether $a_i=b_i$ for all $i=1,\ldots,p$
while the pairwise counterpart evaluates the cardinality of $\{(i,j)|a_i< a_j, b_i< b_j\}$ divided by $p(p-1)/2$.
(The computer was a Laptop-R1DBLO67 Intel(R) Core(TM) i5-8265U CPU
@1.60 GHz 8 GB RAM).

For the binary variables, the previous method \cite{ina11} that searches the order uses the G-test that is asymptotically equivalent to testing by mutual information.
We assume that the conventional procedure uses mutual information
estimates for the independence test when comparing it with the proposed procedure.

\subsection{When no confounding exists}

When no confounding exists, we generated $n$ examples such that
$X_i=\sum_{j=1}^{i-1}X_j +e_i$ for $i=1,\ldots,p$, where $e_i$ takes value one with a probability that is chosen randomly (uniformly) over $\{0.1,0,2,\ldots,0.9\}$, and the variable size $p$ ranges over $2,4,6,8,10,12$.
We repeated the order identifications one hundred times for each pair of sample and variable sizes $(n,p)$.

From Figure \ref{fig002}, we observe that the proposed method exhibits considerably better performance than the conventional method.
On the other hand, measured by the pairwise criterion, the performance difference is not significant because
the proposed and conventional methods identify the orders globally and locally, respectively (Figure \ref{fig002}).

\subsection{When confounding exists}

Next, we added local and global confounders to the sequences generated by the model without confounding. Specifically, we add correlated noise to two and to more than two variables. 
For the local confounders, in the experiments, we independently flipped each of $X_{2i-1}$ and $X_{2i}$ with probability 0.2 for $i=1,2,\ldots,p/2$, where $p$ ranges over $2,4,6,8,10$.

From Figure \ref{fig5}, we observe that even when confounding exists, the proposed method still performs better, particularly on the complete matching criterion.

In section 4, we proved that the number of mutual information values computed is at most $p(p-1)/2$ out of $p(2^{p-1}-1)$ if no confounding exists. However, even if confounding exists, we see that the computational load is not significantly large (at most 50\% more) compared with the no confounding case (Table \ref{tab3}).

For the global confounders, in the experiments, we independently flipped each of $X_{j}$ with $j\in S=\{1,2,3\}$ with probability 0.2 for $p=4$.
We replaced $S$ with $\{1,3,5,6\}, \{1,3,5,6,8\}$ for $p=6,8$ and continued the experiments.

In Figures \ref{fig5} and \ref{fig4}, similar phenomena are observed:
regardless of local and global confounding, the proposed procedure
outperforms the conventional procedure, and the computation does not increase even when confounding exists.
Note that relatively little computation is required in general, even when confounding exists.

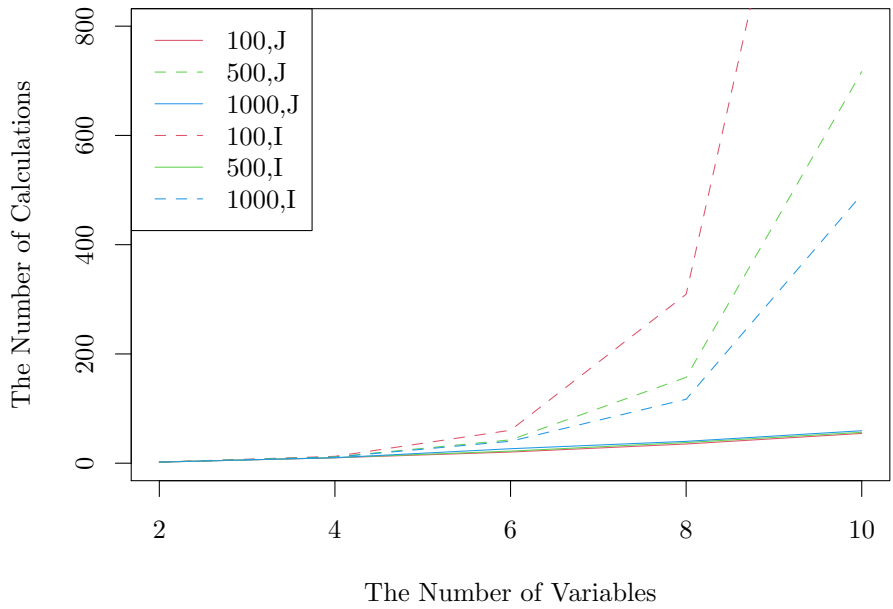
\begin{figure}
    \centering
\begin{tabular}{c}
\input{fig_XX}\\
\input{fig_YY}
\end{tabular}
\caption{$J_n$ requires much less computation than $I_n$, although the
  estimates are close, particularly for large $n$.}
    \label{figXX}
\end{figure}

\subsection{Estimation of Mutual Information}
We have seen that estimates $I_n$ and $J_n$ take larger and exact values, respectively, although they converge to the true as $n$ grows. 

For simplicity, suppose that no confounding exists. 
Then, for $I_n$, the nodes closer to the start tend to be chosen among the open nodes in Algorithm 1 because the sum of the mutual information of the variables in the true order is estimated to be positive. However, $J_n$ does not overfit, and the variable in the true order tends to be chosen from the open variables.

As we have seen in Figure \ref{figXX}, $J_n$ requires much less computation than $I_n$, although both of them estimate correctly for large $n$.

\subsection{The Asia Dataset}
Asia is a well-known dataset. Lauritzen and Spiegelhalter (1988) stated the following:
"Shortness-of-breath (dyspnoea) may be due to tuberculosis, 
lung cancer or bronchitis, or none of them, or more than one of them. 
A recent visit to Asia increases the chances of tuberculosis, while smoking is known to be a risk factor for both lung cancer and bronchitis. 
The results of a single chest X-ray do not discriminate between lung cancer and tuberculosis, as neither does the presence or absence of dyspnoea."
The data set consists of $n=5000$ observations and $p=8$ variables.

From the dataset and proposed algorithm, we obtain the following order:
visit to Asia $\rightarrow$ tuberculosis $\rightarrow$ chest X-ray $\rightarrow$ lung cancer $\rightarrow$ tuberculosis or lung cancer
$\rightarrow$ dyspnoea $\rightarrow$ bronchitis $\rightarrow$
smoking.

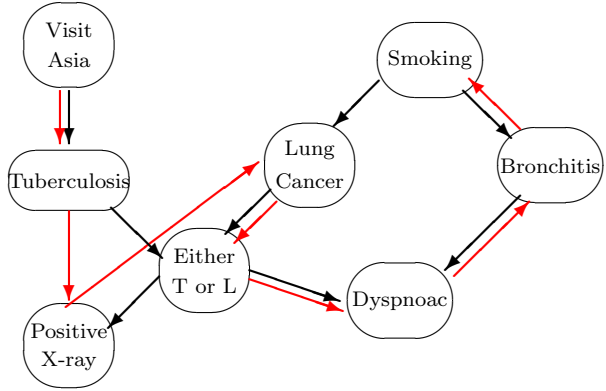
\begin{figure*}
\input{figure090}
\caption{The binary LiNGAM orders the eight variables (the seven red arrows).
The order seems to be  consistent with the BN except for the V structure at Dyspnoac, while the criteria of 
BN structure learning and LiNGAM are different.
\label{figure090}}
\end{figure*}

Bayesian network structure learning (BNSL) procedures construct different structures,  each of which suggests the causal order because the criteria are different. 
Although the order for the  Asia model is consistent with the original one except for the V structure at Dyspnoac (Figure \ref{figure090}),
we may consider choosing the best of the Markov equivalent BN structures using the LiNGAM.
However, the BNSL cannot identify the causal order when more than one Markov-equivalent structure exists.
The proposed method is helpful because we are not concerned with whether confounding exists when inferring causality.

\section{Concluding Remarks}

We proposed how to quantitatively evaluate confounding via mutual information as well as how to obtain the causal order among the variables that minimizes the mutual information.
Thus far, LiNGAM dealt with only the case in which variables are free from confounding, which was very restrictive.
This paper formalizes so that minimizing the mutual information is choosing the causal order with the least confounding.
The original LiNGAM seeks the causal order only for the particular case in which the true model has zero mutual information.

In particular, we formulated LiNGAM based on minimizing the mutual information value and making LiNGAM available even when confounding is present,
reduced finding the optimal causal order to the shortest path problem,
applied mutual information estimation based on the Bayes/MDL criteria, and
finally showed that the proposed LiNGAM achieves significantly better performance.

Forthcoming work will include the case in which the variables are continuous.
Also, the scalability, for how large $p$ the LiNGAM works,  may be of interest. 
This paper focuses on the correctness and does not claim that it works for large $p$.
In fact, the size $p$ is not too large when examining the variable order using the LiNGAM.
When we extend the framework to the continuous variables, we would examine the scalability.

\bibliography{2018-2-7}

\end{document}

%% file: figure007.tex
\setlength\unitlength{0.35mm}
\small
\begin{center}
\begin{tabular}{cccc}
\begin{picture}(120,80)(0,20)
\put(10,95){(a)}
\put(40,90){\framebox(40,10){$\{X,Y,Z\}$}}
\put(60,90){\line(-4,-1){40}}
\put(60,90){\line(0,-1){10}}
\put(60,90){\line(4,-1){40}}
\put(5,70){\color{blue}\framebox(30,10){$\{Y,Z\}$}}
\put(45,70){\color{blue}\framebox(30,10){$\{Z,X\}$}}
\put(85,70){\color{blue}\framebox(30,10){$\{X,Y\}$}}
\put(60,70){\line(-2,-1){40}}
\put(60,70){\line(2,-1){40}}
\put(20,70){\line(2,-1){40}}
\put(100,70){\line(-2,-1){40}}
\put(20,70){\line(0,-1){20}}
\put(100,70){\line(0,-1){20}}
\put(10,40){\framebox(20,10){$\{Z\}$}}
\put(50,40){\framebox(20,10){$\{Y\}$}}
\put(90,40){\framebox(20,10){$\{X\}$}}
\put(20,40){\line(4,-1){40}}
\put(100,40){\line(-4,-1){40}}
\put(60,40){\line(0,-1){10}}
\put(50,20){\framebox(20,10){$\{\}$}}
\end{picture}&
\begin{picture}(120,80)(0,20)
\put(10,95){(b)}
\put(40,90){\framebox(40,10){$\{X,Y,Z\}$}}
\put(60,90){\line(-4,-1){40}}
\put(60,90){\line(0,-1){10}}
\put(60,90){\line(4,-1){40}}
\put(5,70){\framebox(30,10){$\{Y,Z\}$}}
\put(45,70){\color{blue}\framebox(30,10){$\{Z,X\}$}}
\put(85,70){\color{blue}\framebox(30,10){$\{X,Y\}$}}
\put(60,70){\line(-2,-1){40}}
\put(60,70){\line(2,-1){40}}
\put(20,70){\line(2,-1){40}}
\put(100,70){\line(-2,-1){40}}
\put(20,70){\line(0,-1){20}}
\put(100,70){\line(0,-1){20}}
\put(10,40){\color{blue}\framebox(20,10){$\{Z\}$}}
\put(50,40){\color{blue}\framebox(20,10){$\{Y\}$}}
\put(90,40){\framebox(20,10){$\{X\}$}}
\put(20,40){\line(4,-1){40}}
\put(100,40){\line(-4,-1){40}}
\put(60,40){\line(0,-1){10}}
\put(50,20){\framebox(20,10){$\{\}$}}\end{picture}&
\begin{picture}(120,80)(0,20)
\put(10,95){(c)}
\put(40,90){\framebox(40,10){$\{X,Y,Z\}$}}
\put(60,90){\line(0,-1){10}}
\put(60,90){\line(4,-1){40}}
\put(5,70){\framebox(30,10){$\{Y,Z\}$}}
\put(45,70){\color{blue}\framebox(30,10){$\{Z,X\}$}}
\put(85,70){\color{blue}\framebox(30,10){$\{X,Y\}$}}
\put(60,70){\line(-2,-1){40}}
\put(60,70){\line(2,-1){40}}
\put(20,70){\line(2,-1){40}}
\put(100,70){\line(-2,-1){40}}
\put(100,70){\line(0,-1){20}}
\put(10,40){\framebox(20,10){$\{Z\}$}}
\put(50,40){\color{blue}\framebox(20,10){$\{Y\}$}}
\put(90,40){\framebox(20,10){$\{X\}$}}
\put(100,40){\line(-4,-1){40}}
\put(60,40){\line(0,-1){10}}
\put(50,20){\color{red}\framebox(20,10){$\{\}$}}
\put(20,40){\color{red}\line(4,-1){40}}
\put(20,70){\color{red}\line(0,-1){20}}
\put(60,90){\color{red}\line(-4,-1){40}}
\end{picture}\\
\begin{picture}(120,80)(0,30)
\put(10,95){(d)}
\put(40,90){\framebox(40,10){$\{X,Y,Z\}$}}
\put(60,90){\line(0,-1){10}}
\put(60,90){\line(4,-1){40}}
\put(5,70){\framebox(30,10){$\{Y,Z\}$}}
\put(45,70){\color{blue}\framebox(30,10){$\{Z,X\}$}}
\put(85,70){\color{blue}\framebox(30,10){$\{X,Y\}$}}
\put(60,70){\line(-2,-1){40}}
\put(60,70){\line(2,-1){40}}
\put(20,70){\color{red}\line(2,-1){40}}
\put(100,70){\line(-2,-1){40}}
\put(100,70){\line(0,-1){20}}
\put(10,40){\color{blue}\framebox(20,10){$\{Z\}$}}
\put(50,40){\framebox(20,10){$\{Y\}$}}
\put(90,40){\framebox(20,10){$\{X\}$}}
\put(100,40){\line(-4,-1){40}}
\put(20,40){\line(4,-1){40}}
\put(60,40){\color{red}\line(0,-1){10}}
\put(50,20){\color{red}\framebox(20,10){$\{\}$}}
\put(20,70){\line(0,-1){20}}
\put(60,90){\color{red}\line(-4,-1){40}}
\end{picture}&
\begin{picture}(120,80)(0,30)
\put(10,95){(e)}
\put(40,90){\framebox(40,10){$\{X,Y,Z\}$}}
\put(60,90){\line(0,-1){10}}
\put(60,90){\line(4,-1){40}}
\put(5,70){\framebox(30,10){$\{Y,Z\}$}}
\put(45,70){\framebox(30,10){$\{Z,X\}$}}
\put(85,70){\color{blue}\framebox(30,10){$\{X,Y\}$}}
\put(60,70){\line(-2,-1){40}}
\put(60,70){\line(2,-1){40}}
\put(20,70){\line(2,-1){40}}
\put(100,70){\line(-2,-1){40}}
\put(100,70){\line(0,-1){20}}
\put(10,40){\color{blue}\framebox(20,10){$\{Z\}$}}
\put(50,40){\color{blue}\framebox(20,10){$\{Y\}$}}
\put(90,40){\color{blue}\framebox(20,10){$\{X\}$}}
\put(100,40){\line(-4,-1){40}}
\put(20,40){\line(4,-1){40}}
\put(60,40){\line(0,-1){10}}
\put(50,20){\framebox(20,10){$\{\}$}}\put(20,70){\line(0,-1){20}}
\put(60,90){\line(-4,-1){40}}
\end{picture}&
\begin{picture}(120,80)(0,30)
\put(10,95){(f)}
\put(40,90){\framebox(40,10){$\{X,Y,Z\}$}}
\put(60,90){\line(0,-1){10}}
\put(60,90){\line(4,-1){40}}
\put(5,70){\framebox(30,10){$\{Y,Z\}$}}
\put(45,70){\color{blue}\framebox(30,10){$\{Z,X\}$}}
\put(85,70){\framebox(30,10){$\{X,Y\}$}}
\put(60,70){\line(-2,-1){40}}
\put(60,70){\line(2,-1){40}}
\put(20,70){\line(2,-1){40}}
\put(100,70){\line(-2,-1){40}}
\put(100,70){\line(0,-1){20}}
\put(10,40){\color{blue}\framebox(20,10){$\{Z\}$}}
\put(50,40){\color{blue}\framebox(20,10){$\{Y\}$}}
\put(90,40){\color{blue}\framebox(20,10){$\{X\}$}}
\put(100,40){\line(-4,-1){40}}
\put(20,40){\line(4,-1){40}}
\put(60,40){\line(0,-1){10}}
\put(50,20){\framebox(20,10){$\{\}$}}\put(20,70){\line(0,-1){20}}
\put(60,90){\line(-4,-1){40}}
\end{picture}
\end{tabular}
\end{center}

%% file: nc_rate1.tex
\begin{tikzpicture}[x=0.60pt,y=0.45pt]
\definecolor{fillColor}{RGB}{255,255,255}
\path[use as bounding box,fill=fillColor,fill opacity=0.00] (0,0) rectangle (361.35,361.35);
\begin{scope}
\path[clip] (  0.00,  0.00) rectangle (361.35,361.35);
\definecolor{drawColor}{RGB}{0,0,0}

\path[draw=drawColor,line width= 0.4pt,line join=round,line cap=round] ( 59.83, 61.20) -- (325.52, 61.20);

\path[draw=drawColor,line width= 0.4pt,line join=round,line cap=round] ( 59.83, 61.20) -- ( 59.83, 55.20);

\path[draw=drawColor,line width= 0.4pt,line join=round,line cap=round] (112.97, 61.20) -- (112.97, 55.20);

\path[draw=drawColor,line width= 0.4pt,line join=round,line cap=round] (166.11, 61.20) -- (166.11, 55.20);

\path[draw=drawColor,line width= 0.4pt,line join=round,line cap=round] (219.24, 61.20) -- (219.24, 55.20);

\path[draw=drawColor,line width= 0.4pt,line join=round,line cap=round] (272.38, 61.20) -- (272.38, 55.20);

\path[draw=drawColor,line width= 0.4pt,line join=round,line cap=round] (325.52, 61.20) -- (325.52, 55.20);

\node[text=drawColor,anchor=base,inner sep=0pt, outer sep=0pt, scale=  1.00] at ( 59.83, 39.60) {2};

\node[text=drawColor,anchor=base,inner sep=0pt, outer sep=0pt, scale=  1.00] at (112.97, 39.60) {4};

\node[text=drawColor,anchor=base,inner sep=0pt, outer sep=0pt, scale=  1.00] at (166.11, 39.60) {6};

\node[text=drawColor,anchor=base,inner sep=0pt, outer sep=0pt, scale=  1.00] at (219.24, 39.60) {8};

\node[text=drawColor,anchor=base,inner sep=0pt, outer sep=0pt, scale=  1.00] at (272.38, 39.60) {10};

\node[text=drawColor,anchor=base,inner sep=0pt, outer sep=0pt, scale=  1.00] at (325.52, 39.60) {12};

\path[draw=drawColor,line width= 0.4pt,line join=round,line cap=round] ( 49.20, 70.49) -- ( 49.20,302.86);

\path[draw=drawColor,line width= 0.4pt,line join=round,line cap=round] ( 49.20, 70.49) -- ( 43.20, 70.49);

\path[draw=drawColor,line width= 0.4pt,line join=round,line cap=round] ( 49.20,116.97) -- ( 43.20,116.97);

\path[draw=drawColor,line width= 0.4pt,line join=round,line cap=round] ( 49.20,163.44) -- ( 43.20,163.44);

\path[draw=drawColor,line width= 0.4pt,line join=round,line cap=round] ( 49.20,209.91) -- ( 43.20,209.91);

\path[draw=drawColor,line width= 0.4pt,line join=round,line cap=round] ( 49.20,256.38) -- ( 43.20,256.38);

\path[draw=drawColor,line width= 0.4pt,line join=round,line cap=round] ( 49.20,302.86) -- ( 43.20,302.86);

\node[text=drawColor,rotate= 90.00,anchor=base,inner sep=0pt, outer sep=0pt, scale=  1.00] at ( 34.80, 70.49) {0.0};

\node[text=drawColor,rotate= 90.00,anchor=base,inner sep=0pt, outer sep=0pt, scale=  1.00] at ( 34.80,116.97) {0.2};

\node[text=drawColor,rotate= 90.00,anchor=base,inner sep=0pt, outer sep=0pt, scale=  1.00] at ( 34.80,163.44) {0.4};

\node[text=drawColor,rotate= 90.00,anchor=base,inner sep=0pt, outer sep=0pt, scale=  1.00] at ( 34.80,209.91) {0.6};

\node[text=drawColor,rotate= 90.00,anchor=base,inner sep=0pt, outer sep=0pt, scale=  1.00] at ( 34.80,256.38) {0.8};

\node[text=drawColor,rotate= 90.00,anchor=base,inner sep=0pt, outer sep=0pt, scale=  1.00] at ( 34.80,302.86) {1.0};

\path[draw=drawColor,line width= 0.4pt,line join=round,line cap=round] ( 49.20, 61.20) --
	(336.15, 61.20) --
	(336.15,312.15) --
	( 49.20,312.15) --
	( 49.20, 61.20);
\end{scope}
\begin{scope}
\path[clip] (  0.00,  0.00) rectangle (361.35,361.35);
\definecolor{drawColor}{RGB}{0,0,0}

\node[text=drawColor,anchor=base,inner sep=0pt, outer sep=0pt, scale=  1.20] at (192.68,332.61) {\bfseries No Confounding (Complete)};

\node[text=drawColor,anchor=base,inner sep=0pt, outer sep=0pt, scale=  1.00] at (192.68, 15.60) {The Number of  Variables p};

\node[text=drawColor,rotate= 90.00,anchor=base,inner sep=0pt, outer sep=0pt, scale=  1.00] at ( 10.80,186.67) {Correct Rate};
\end{scope}
\begin{scope}
\path[clip] ( 49.20, 61.20) rectangle (336.15,312.15);
\definecolor{drawColor}{RGB}{255,0,0}

\path[draw=drawColor,line width= 0.4pt,line join=round,line cap=round] ( 59.83,286.59) --
	(112.97,251.74) --
	(166.11,235.47) --
	(219.24,177.38) --
	(272.38,154.14) --
	(325.52,147.17);
\definecolor{drawColor}{RGB}{0,0,0}

\path[draw=drawColor,line width= 0.4pt,line join=round,line cap=round] ( 59.83,290.09) --
	( 62.86,284.84) --
	( 56.80,284.84) --
	( 59.83,290.09);

\path[draw=drawColor,line width= 0.4pt,line join=round,line cap=round] (112.97,255.24) --
	(116.00,249.99) --
	(109.94,249.99) --
	(112.97,255.24);

\path[draw=drawColor,line width= 0.4pt,line join=round,line cap=round] (166.11,238.97) --
	(169.14,233.72) --
	(163.08,233.72) --
	(166.11,238.97);

\path[draw=drawColor,line width= 0.4pt,line join=round,line cap=round] (219.24,180.88) --
	(222.27,175.63) --
	(216.21,175.63) --
	(219.24,180.88);

\path[draw=drawColor,line width= 0.4pt,line join=round,line cap=round] (272.38,157.64) --
	(275.41,152.39) --
	(269.35,152.39) --
	(272.38,157.64);

\path[draw=drawColor,line width= 0.4pt,line join=round,line cap=round] (325.52,150.67) --
	(328.55,145.42) --
	(322.49,145.42) --
	(325.52,150.67);
\definecolor{drawColor}{RGB}{0,0,255}

\path[draw=drawColor,line width= 0.4pt,line join=round,line cap=round] ( 59.83,302.86) --
	(112.97,300.53) --
	(166.11,295.88) --
	(219.24,291.24) --
	(272.38,291.24) --
	(325.52,284.27);
\definecolor{drawColor}{RGB}{0,0,0}

\path[draw=drawColor,line width= 0.4pt,line join=round,line cap=round] ( 56.65,302.86) -- ( 63.01,302.86);

\path[draw=drawColor,line width= 0.4pt,line join=round,line cap=round] ( 59.83,299.67) -- ( 59.83,306.04);

\path[draw=drawColor,line width= 0.4pt,line join=round,line cap=round] (109.78,300.53) -- (116.15,300.53);

\path[draw=drawColor,line width= 0.4pt,line join=round,line cap=round] (112.97,297.35) -- (112.97,303.71);

\path[draw=drawColor,line width= 0.4pt,line join=round,line cap=round] (162.92,295.88) -- (169.29,295.88);

\path[draw=drawColor,line width= 0.4pt,line join=round,line cap=round] (166.11,292.70) -- (166.11,299.07);

\path[draw=drawColor,line width= 0.4pt,line join=round,line cap=round] (216.06,291.24) -- (222.43,291.24);

\path[draw=drawColor,line width= 0.4pt,line join=round,line cap=round] (219.24,288.06) -- (219.24,294.42);

\path[draw=drawColor,line width= 0.4pt,line join=round,line cap=round] (269.20,291.24) -- (275.57,291.24);

\path[draw=drawColor,line width= 0.4pt,line join=round,line cap=round] (272.38,288.06) -- (272.38,294.42);

\path[draw=drawColor,line width= 0.4pt,line join=round,line cap=round] (322.34,284.27) -- (328.70,284.27);

\path[draw=drawColor,line width= 0.4pt,line join=round,line cap=round] (325.52,281.08) -- (325.52,287.45);
\definecolor{drawColor}{RGB}{0,255,0}

\path[draw=drawColor,line width= 0.4pt,line join=round,line cap=round] ( 59.83,302.86) --
	(112.97,302.86) --
	(166.11,302.86) --
	(219.24,302.86) --
	(272.38,302.86) --
	(325.52,293.56);
\definecolor{drawColor}{RGB}{0,0,0}

\path[draw=drawColor,line width= 0.4pt,line join=round,line cap=round] ( 57.58,300.61) -- ( 62.08,305.11);

\path[draw=drawColor,line width= 0.4pt,line join=round,line cap=round] ( 57.58,305.11) -- ( 62.08,300.61);

\path[draw=drawColor,line width= 0.4pt,line join=round,line cap=round] (110.72,300.61) -- (115.22,305.11);

\path[draw=drawColor,line width= 0.4pt,line join=round,line cap=round] (110.72,305.11) -- (115.22,300.61);

\path[draw=drawColor,line width= 0.4pt,line join=round,line cap=round] (163.86,300.61) -- (168.36,305.11);

\path[draw=drawColor,line width= 0.4pt,line join=round,line cap=round] (163.86,305.11) -- (168.36,300.61);

\path[draw=drawColor,line width= 0.4pt,line join=round,line cap=round] (216.99,300.61) -- (221.49,305.11);

\path[draw=drawColor,line width= 0.4pt,line join=round,line cap=round] (216.99,305.11) -- (221.49,300.61);

\path[draw=drawColor,line width= 0.4pt,line join=round,line cap=round] (270.13,300.61) -- (274.63,305.11);

\path[draw=drawColor,line width= 0.4pt,line join=round,line cap=round] (270.13,305.11) -- (274.63,300.61);

\path[draw=drawColor,line width= 0.4pt,line join=round,line cap=round] (323.27,291.31) -- (327.77,295.81);

\path[draw=drawColor,line width= 0.4pt,line join=round,line cap=round] (323.27,295.81) -- (327.77,291.31);
\definecolor{drawColor}{RGB}{255,0,0}

\path[draw=drawColor,line width= 0.4pt,dash pattern=on 1pt off 3pt ,line join=round,line cap=round] ( 59.83,291.24) --
	(112.97,177.38) --
	(166.11, 98.38) --
	(219.24, 72.82) --
	(272.38, 70.49) --
	(325.52, 70.49);
\definecolor{drawColor}{RGB}{0,0,0}

\path[draw=drawColor,line width= 0.4pt,line join=round,line cap=round] ( 56.65,291.24) --
	( 59.83,294.42) --
	( 63.01,291.24) --
	( 59.83,288.06) --
	( 56.65,291.24);

\path[draw=drawColor,line width= 0.4pt,line join=round,line cap=round] (109.78,177.38) --
	(112.97,180.56) --
	(116.15,177.38) --
	(112.97,174.20) --
	(109.78,177.38);

\path[draw=drawColor,line width= 0.4pt,line join=round,line cap=round] (162.92, 98.38) --
	(166.11,101.56) --
	(169.29, 98.38) --
	(166.11, 95.20) --
	(162.92, 98.38);

\path[draw=drawColor,line width= 0.4pt,line join=round,line cap=round] (216.06, 72.82) --
	(219.24, 76.00) --
	(222.43, 72.82) --
	(219.24, 69.64) --
	(216.06, 72.82);

\path[draw=drawColor,line width= 0.4pt,line join=round,line cap=round] (269.20, 70.49) --
	(272.38, 73.68) --
	(275.57, 70.49) --
	(272.38, 67.31) --
	(269.20, 70.49);

\path[draw=drawColor,line width= 0.4pt,line join=round,line cap=round] (322.34, 70.49) --
	(325.52, 73.68) --
	(328.70, 70.49) --
	(325.52, 67.31) --
	(322.34, 70.49);
\definecolor{drawColor}{RGB}{0,0,255}

\path[draw=drawColor,line width= 0.4pt,dash pattern=on 1pt off 3pt ,line join=round,line cap=round] ( 59.83,300.53) --
	(112.97,242.44) --
	(166.11,119.29) --
	(219.24, 75.14) --
	(272.38, 75.14) --
	(325.52, 70.49);
\definecolor{drawColor}{RGB}{0,0,0}

\path[draw=drawColor,line width= 0.4pt,line join=round,line cap=round] ( 59.83,297.03) --
	( 62.86,302.28) --
	( 56.80,302.28) --
	( 59.83,297.03);

\path[draw=drawColor,line width= 0.4pt,line join=round,line cap=round] (112.97,238.94) --
	(116.00,244.19) --
	(109.94,244.19) --
	(112.97,238.94);

\path[draw=drawColor,line width= 0.4pt,line join=round,line cap=round] (166.11,115.79) --
	(169.14,121.04) --
	(163.08,121.04) --
	(166.11,115.79);

\path[draw=drawColor,line width= 0.4pt,line join=round,line cap=round] (219.24, 71.64) --
	(222.27, 76.89) --
	(216.21, 76.89) --
	(219.24, 71.64);

\path[draw=drawColor,line width= 0.4pt,line join=round,line cap=round] (272.38, 71.64) --
	(275.41, 76.89) --
	(269.35, 76.89) --
	(272.38, 71.64);

\path[draw=drawColor,line width= 0.4pt,line join=round,line cap=round] (325.52, 67.00) --
	(328.55, 72.24) --
	(322.49, 72.24) --
	(325.52, 67.00);
\definecolor{drawColor}{RGB}{0,255,0}

\path[draw=drawColor,line width= 0.4pt,dash pattern=on 1pt off 3pt ,line join=round,line cap=round] ( 59.83,300.53) --
	(112.97,270.32) --
	(166.11,172.73) --
	(219.24, 93.73) --
	(272.38, 72.82) --
	(325.52, 70.49);
\definecolor{drawColor}{RGB}{0,0,0}

\path[draw=drawColor,line width= 0.4pt,line join=round,line cap=round] ( 57.58,298.28) rectangle ( 62.08,302.78);

\path[draw=drawColor,line width= 0.4pt,line join=round,line cap=round] ( 57.58,298.28) -- ( 62.08,302.78);

\path[draw=drawColor,line width= 0.4pt,line join=round,line cap=round] ( 57.58,302.78) -- ( 62.08,298.28);

\path[draw=drawColor,line width= 0.4pt,line join=round,line cap=round] (110.72,268.07) rectangle (115.22,272.57);

\path[draw=drawColor,line width= 0.4pt,line join=round,line cap=round] (110.72,268.07) -- (115.22,272.57);

\path[draw=drawColor,line width= 0.4pt,line join=round,line cap=round] (110.72,272.57) -- (115.22,268.07);

\path[draw=drawColor,line width= 0.4pt,line join=round,line cap=round] (163.86,170.48) rectangle (168.36,174.98);

\path[draw=drawColor,line width= 0.4pt,line join=round,line cap=round] (163.86,170.48) -- (168.36,174.98);

\path[draw=drawColor,line width= 0.4pt,line join=round,line cap=round] (163.86,174.98) -- (168.36,170.48);

\path[draw=drawColor,line width= 0.4pt,line join=round,line cap=round] (216.99, 91.48) rectangle (221.49, 95.98);

\path[draw=drawColor,line width= 0.4pt,line join=round,line cap=round] (216.99, 91.48) -- (221.49, 95.98);

\path[draw=drawColor,line width= 0.4pt,line join=round,line cap=round] (216.99, 95.98) -- (221.49, 91.48);

\path[draw=drawColor,line width= 0.4pt,line join=round,line cap=round] (270.13, 70.57) rectangle (274.63, 75.07);

\path[draw=drawColor,line width= 0.4pt,line join=round,line cap=round] (270.13, 70.57) -- (274.63, 75.07);

\path[draw=drawColor,line width= 0.4pt,line join=round,line cap=round] (270.13, 75.07) -- (274.63, 70.57);

\path[draw=drawColor,line width= 0.4pt,line join=round,line cap=round] (323.27, 68.24) rectangle (327.77, 72.74);

\path[draw=drawColor,line width= 0.4pt,line join=round,line cap=round] (323.27, 68.24) -- (327.77, 72.74);

\path[draw=drawColor,line width= 0.4pt,line join=round,line cap=round] (323.27, 72.74) -- (327.77, 68.24);
\end{scope}
\end{tikzpicture}

%% file: nc_rate2.tex
\begin{tikzpicture}[x=0.60pt,y=0.45pt]
\definecolor{fillColor}{RGB}{255,255,255}
\path[use as bounding box,fill=fillColor,fill opacity=0.00] (0,0) rectangle (361.35,361.35);
\begin{scope}
\path[clip] (  0.00,  0.00) rectangle (361.35,361.35);
\definecolor{drawColor}{RGB}{0,0,0}

\path[draw=drawColor,line width= 0.4pt,line join=round,line cap=round] ( 59.83, 61.20) -- (325.52, 61.20);

\path[draw=drawColor,line width= 0.4pt,line join=round,line cap=round] ( 59.83, 61.20) -- ( 59.83, 55.20);

\path[draw=drawColor,line width= 0.4pt,line join=round,line cap=round] (112.97, 61.20) -- (112.97, 55.20);

\path[draw=drawColor,line width= 0.4pt,line join=round,line cap=round] (166.11, 61.20) -- (166.11, 55.20);

\path[draw=drawColor,line width= 0.4pt,line join=round,line cap=round] (219.24, 61.20) -- (219.24, 55.20);

\path[draw=drawColor,line width= 0.4pt,line join=round,line cap=round] (272.38, 61.20) -- (272.38, 55.20);

\path[draw=drawColor,line width= 0.4pt,line join=round,line cap=round] (325.52, 61.20) -- (325.52, 55.20);

\node[text=drawColor,anchor=base,inner sep=0pt, outer sep=0pt, scale=  1.00] at ( 59.83, 39.60) {2};

\node[text=drawColor,anchor=base,inner sep=0pt, outer sep=0pt, scale=  1.00] at (112.97, 39.60) {4};

\node[text=drawColor,anchor=base,inner sep=0pt, outer sep=0pt, scale=  1.00] at (166.11, 39.60) {6};

\node[text=drawColor,anchor=base,inner sep=0pt, outer sep=0pt, scale=  1.00] at (219.24, 39.60) {8};

\node[text=drawColor,anchor=base,inner sep=0pt, outer sep=0pt, scale=  1.00] at (272.38, 39.60) {10};

\node[text=drawColor,anchor=base,inner sep=0pt, outer sep=0pt, scale=  1.00] at (325.52, 39.60) {12};

\path[draw=drawColor,line width= 0.4pt,line join=round,line cap=round] ( 49.20, 70.49) -- ( 49.20,302.86);

\path[draw=drawColor,line width= 0.4pt,line join=round,line cap=round] ( 49.20, 70.49) -- ( 43.20, 70.49);

\path[draw=drawColor,line width= 0.4pt,line join=round,line cap=round] ( 49.20,116.97) -- ( 43.20,116.97);

\path[draw=drawColor,line width= 0.4pt,line join=round,line cap=round] ( 49.20,163.44) -- ( 43.20,163.44);

\path[draw=drawColor,line width= 0.4pt,line join=round,line cap=round] ( 49.20,209.91) -- ( 43.20,209.91);

\path[draw=drawColor,line width= 0.4pt,line join=round,line cap=round] ( 49.20,256.38) -- ( 43.20,256.38);

\path[draw=drawColor,line width= 0.4pt,line join=round,line cap=round] ( 49.20,302.86) -- ( 43.20,302.86);

\node[text=drawColor,rotate= 90.00,anchor=base,inner sep=0pt, outer sep=0pt, scale=  1.00] at ( 34.80, 70.49) {0.0};

\node[text=drawColor,rotate= 90.00,anchor=base,inner sep=0pt, outer sep=0pt, scale=  1.00] at ( 34.80,116.97) {0.2};

\node[text=drawColor,rotate= 90.00,anchor=base,inner sep=0pt, outer sep=0pt, scale=  1.00] at ( 34.80,163.44) {0.4};

\node[text=drawColor,rotate= 90.00,anchor=base,inner sep=0pt, outer sep=0pt, scale=  1.00] at ( 34.80,209.91) {0.6};

\node[text=drawColor,rotate= 90.00,anchor=base,inner sep=0pt, outer sep=0pt, scale=  1.00] at ( 34.80,256.38) {0.8};

\node[text=drawColor,rotate= 90.00,anchor=base,inner sep=0pt, outer sep=0pt, scale=  1.00] at ( 34.80,302.86) {1.0};

\path[draw=drawColor,line width= 0.4pt,line join=round,line cap=round] ( 49.20, 61.20) --
	(336.15, 61.20) --
	(336.15,312.15) --
	( 49.20,312.15) --
	( 49.20, 61.20);
\end{scope}
\begin{scope}
\path[clip] (  0.00,  0.00) rectangle (361.35,361.35);
\definecolor{drawColor}{RGB}{0,0,0}

\node[text=drawColor,anchor=base,inner sep=0pt, outer sep=0pt, scale=  1.20] at (192.68,332.61) {\bfseries No Confounding (Pairwise)};

\node[text=drawColor,anchor=base,inner sep=0pt, outer sep=0pt, scale=  1.00] at (192.68, 15.60) {The Number of  Variables p};

\node[text=drawColor,rotate= 90.00,anchor=base,inner sep=0pt, outer sep=0pt, scale=  1.00] at ( 10.80,186.67) {Correct Rate};
\end{scope}
\begin{scope}
\path[clip] ( 49.20, 61.20) rectangle (336.15,312.15);
\definecolor{drawColor}{RGB}{255,0,0}

\path[draw=drawColor,line width= 0.4pt,line join=round,line cap=round] ( 59.83,286.59) --
	(112.97,279.62) --
	(166.11,287.83) --
	(219.24,274.23) --
	(272.38,257.26) --
	(325.52,266.14);
\definecolor{drawColor}{RGB}{0,0,0}

\path[draw=drawColor,line width= 0.4pt,line join=round,line cap=round] ( 59.83,290.09) --
	( 62.86,284.84) --
	( 56.80,284.84) --
	( 59.83,290.09);

\path[draw=drawColor,line width= 0.4pt,line join=round,line cap=round] (112.97,283.12) --
	(116.00,277.87) --
	(109.94,277.87) --
	(112.97,283.12);

\path[draw=drawColor,line width= 0.4pt,line join=round,line cap=round] (166.11,291.33) --
	(169.14,286.08) --
	(163.08,286.08) --
	(166.11,291.33);

\path[draw=drawColor,line width= 0.4pt,line join=round,line cap=round] (219.24,277.72) --
	(222.27,272.48) --
	(216.21,272.48) --
	(219.24,277.72);

\path[draw=drawColor,line width= 0.4pt,line join=round,line cap=round] (272.38,260.76) --
	(275.41,255.51) --
	(269.35,255.51) --
	(272.38,260.76);

\path[draw=drawColor,line width= 0.4pt,line join=round,line cap=round] (325.52,269.63) --
	(328.55,264.39) --
	(322.49,264.39) --
	(325.52,269.63);
\definecolor{drawColor}{RGB}{0,0,255}

\path[draw=drawColor,line width= 0.4pt,line join=round,line cap=round] ( 59.83,302.86) --
	(112.97,301.69) --
	(166.11,299.45) --
	(219.24,301.11) --
	(272.38,301.25) --
	(325.52,299.48);
\definecolor{drawColor}{RGB}{0,0,0}

\path[draw=drawColor,line width= 0.4pt,line join=round,line cap=round] ( 56.65,302.86) -- ( 63.01,302.86);

\path[draw=drawColor,line width= 0.4pt,line join=round,line cap=round] ( 59.83,299.67) -- ( 59.83,306.04);

\path[draw=drawColor,line width= 0.4pt,line join=round,line cap=round] (109.78,301.69) -- (116.15,301.69);

\path[draw=drawColor,line width= 0.4pt,line join=round,line cap=round] (112.97,298.51) -- (112.97,304.88);

\path[draw=drawColor,line width= 0.4pt,line join=round,line cap=round] (162.92,299.45) -- (169.29,299.45);

\path[draw=drawColor,line width= 0.4pt,line join=round,line cap=round] (166.11,296.27) -- (166.11,302.63);

\path[draw=drawColor,line width= 0.4pt,line join=round,line cap=round] (216.06,301.11) -- (222.43,301.11);

\path[draw=drawColor,line width= 0.4pt,line join=round,line cap=round] (219.24,297.93) -- (219.24,304.29);

\path[draw=drawColor,line width= 0.4pt,line join=round,line cap=round] (269.20,301.25) -- (275.57,301.25);

\path[draw=drawColor,line width= 0.4pt,line join=round,line cap=round] (272.38,298.07) -- (272.38,304.44);

\path[draw=drawColor,line width= 0.4pt,line join=round,line cap=round] (322.34,299.48) -- (328.70,299.48);

\path[draw=drawColor,line width= 0.4pt,line join=round,line cap=round] (325.52,296.29) -- (325.52,302.66);
\definecolor{drawColor}{RGB}{0,255,0}

\path[draw=drawColor,line width= 0.4pt,line join=round,line cap=round] ( 59.83,302.86) --
	(112.97,302.86) --
	(166.11,302.86) --
	(219.24,302.86) --
	(272.38,302.86) --
	(325.52,302.71);
\definecolor{drawColor}{RGB}{0,0,0}

\path[draw=drawColor,line width= 0.4pt,line join=round,line cap=round] ( 57.58,300.61) -- ( 62.08,305.11);

\path[draw=drawColor,line width= 0.4pt,line join=round,line cap=round] ( 57.58,305.11) -- ( 62.08,300.61);

\path[draw=drawColor,line width= 0.4pt,line join=round,line cap=round] (110.72,300.61) -- (115.22,305.11);

\path[draw=drawColor,line width= 0.4pt,line join=round,line cap=round] (110.72,305.11) -- (115.22,300.61);

\path[draw=drawColor,line width= 0.4pt,line join=round,line cap=round] (163.86,300.61) -- (168.36,305.11);

\path[draw=drawColor,line width= 0.4pt,line join=round,line cap=round] (163.86,305.11) -- (168.36,300.61);

\path[draw=drawColor,line width= 0.4pt,line join=round,line cap=round] (216.99,300.61) -- (221.49,305.11);

\path[draw=drawColor,line width= 0.4pt,line join=round,line cap=round] (216.99,305.11) -- (221.49,300.61);

\path[draw=drawColor,line width= 0.4pt,line join=round,line cap=round] (270.13,300.61) -- (274.63,305.11);

\path[draw=drawColor,line width= 0.4pt,line join=round,line cap=round] (270.13,305.11) -- (274.63,300.61);

\path[draw=drawColor,line width= 0.4pt,line join=round,line cap=round] (323.27,300.46) -- (327.77,304.96);

\path[draw=drawColor,line width= 0.4pt,line join=round,line cap=round] (323.27,304.96) -- (327.77,300.46);
\definecolor{drawColor}{RGB}{255,0,0}

\path[draw=drawColor,line width= 0.4pt,dash pattern=on 1pt off 3pt ,line join=round,line cap=round] ( 59.83,291.24) --
	(112.97,270.71) --
	(166.11,260.72) --
	(219.24,241.20) --
	(272.38,234.44) --
	(325.52,232.02);
\definecolor{drawColor}{RGB}{0,0,0}

\path[draw=drawColor,line width= 0.4pt,line join=round,line cap=round] ( 56.65,291.24) --
	( 59.83,294.42) --
	( 63.01,291.24) --
	( 59.83,288.06) --
	( 56.65,291.24);

\path[draw=drawColor,line width= 0.4pt,line join=round,line cap=round] (109.78,270.71) --
	(112.97,273.89) --
	(116.15,270.71) --
	(112.97,267.53) --
	(109.78,270.71);

\path[draw=drawColor,line width= 0.4pt,line join=round,line cap=round] (162.92,260.72) --
	(166.11,263.90) --
	(169.29,260.72) --
	(166.11,257.54) --
	(162.92,260.72);

\path[draw=drawColor,line width= 0.4pt,line join=round,line cap=round] (216.06,241.20) --
	(219.24,244.38) --
	(222.43,241.20) --
	(219.24,238.01) --
	(216.06,241.20);

\path[draw=drawColor,line width= 0.4pt,line join=round,line cap=round] (269.20,234.44) --
	(272.38,237.62) --
	(275.57,234.44) --
	(272.38,231.26) --
	(269.20,234.44);

\path[draw=drawColor,line width= 0.4pt,line join=round,line cap=round] (322.34,232.02) --
	(325.52,235.20) --
	(328.70,232.02) --
	(325.52,228.84) --
	(322.34,232.02);
\definecolor{drawColor}{RGB}{0,0,255}

\path[draw=drawColor,line width= 0.4pt,dash pattern=on 1pt off 3pt ,line join=round,line cap=round] ( 59.83,300.53) --
	(112.97,291.24) --
	(166.11,271.10) --
	(219.24,259.20) --
	(272.38,246.93) --
	(325.52,236.35);
\definecolor{drawColor}{RGB}{0,0,0}

\path[draw=drawColor,line width= 0.4pt,line join=round,line cap=round] ( 59.83,297.03) --
	( 62.86,302.28) --
	( 56.80,302.28) --
	( 59.83,297.03);

\path[draw=drawColor,line width= 0.4pt,line join=round,line cap=round] (112.97,287.74) --
	(116.00,292.99) --
	(109.94,292.99) --
	(112.97,287.74);

\path[draw=drawColor,line width= 0.4pt,line join=round,line cap=round] (166.11,267.60) --
	(169.14,272.85) --
	(163.08,272.85) --
	(166.11,267.60);

\path[draw=drawColor,line width= 0.4pt,line join=round,line cap=round] (219.24,255.71) --
	(222.27,260.95) --
	(216.21,260.95) --
	(219.24,255.71);

\path[draw=drawColor,line width= 0.4pt,line join=round,line cap=round] (272.38,243.43) --
	(275.41,248.68) --
	(269.35,248.68) --
	(272.38,243.43);

\path[draw=drawColor,line width= 0.4pt,line join=round,line cap=round] (325.52,232.85) --
	(328.55,238.10) --
	(322.49,238.10) --
	(325.52,232.85);
\definecolor{drawColor}{RGB}{0,255,0}

\path[draw=drawColor,line width= 0.4pt,dash pattern=on 1pt off 3pt ,line join=round,line cap=round] ( 59.83,300.53) --
	(112.97,297.43) --
	(166.11,283.96) --
	(219.24,264.85) --
	(272.38,253.85) --
	(325.52,242.20);
\definecolor{drawColor}{RGB}{0,0,0}

\path[draw=drawColor,line width= 0.4pt,line join=round,line cap=round] ( 57.58,298.28) rectangle ( 62.08,302.78);

\path[draw=drawColor,line width= 0.4pt,line join=round,line cap=round] ( 57.58,298.28) -- ( 62.08,302.78);

\path[draw=drawColor,line width= 0.4pt,line join=round,line cap=round] ( 57.58,302.78) -- ( 62.08,298.28);

\path[draw=drawColor,line width= 0.4pt,line join=round,line cap=round] (110.72,295.18) rectangle (115.22,299.68);

\path[draw=drawColor,line width= 0.4pt,line join=round,line cap=round] (110.72,295.18) -- (115.22,299.68);

\path[draw=drawColor,line width= 0.4pt,line join=round,line cap=round] (110.72,299.68) -- (115.22,295.18);

\path[draw=drawColor,line width= 0.4pt,line join=round,line cap=round] (163.86,281.71) rectangle (168.36,286.21);

\path[draw=drawColor,line width= 0.4pt,line join=round,line cap=round] (163.86,281.71) -- (168.36,286.21);

\path[draw=drawColor,line width= 0.4pt,line join=round,line cap=round] (163.86,286.21) -- (168.36,281.71);

\path[draw=drawColor,line width= 0.4pt,line join=round,line cap=round] (216.99,262.60) rectangle (221.49,267.10);

\path[draw=drawColor,line width= 0.4pt,line join=round,line cap=round] (216.99,262.60) -- (221.49,267.10);

\path[draw=drawColor,line width= 0.4pt,line join=round,line cap=round] (216.99,267.10) -- (221.49,262.60);

\path[draw=drawColor,line width= 0.4pt,line join=round,line cap=round] (270.13,251.60) rectangle (274.63,256.10);

\path[draw=drawColor,line width= 0.4pt,line join=round,line cap=round] (270.13,251.60) -- (274.63,256.10);

\path[draw=drawColor,line width= 0.4pt,line join=round,line cap=round] (270.13,256.10) -- (274.63,251.60);

\path[draw=drawColor,line width= 0.4pt,line join=round,line cap=round] (323.27,239.95) rectangle (327.77,244.45);

\path[draw=drawColor,line width= 0.4pt,line join=round,line cap=round] (323.27,239.95) -- (327.77,244.45);

\path[draw=drawColor,line width= 0.4pt,line join=round,line cap=round] (323.27,244.45) -- (327.77,239.95);
\definecolor{drawColor}{RGB}{255,0,0}

\path[draw=drawColor,line width= 0.4pt,line join=round,line cap=round] ( 58.20,184) -- ( 76.20,184);
\definecolor{drawColor}{RGB}{0,0,255}

\path[draw=drawColor,line width= 0.4pt,line join=round,line cap=round] ( 58.20,164) -- ( 76.20,164);
\definecolor{drawColor}{RGB}{0,255,0}

\path[draw=drawColor,line width= 0.4pt,line join=round,line cap=round] ( 58.20,144) -- ( 76.20,144);
\definecolor{drawColor}{RGB}{255,0,0}

\path[draw=drawColor,line width= 0.4pt,dash pattern=on 1pt off 3pt ,line join=round,line cap=round] ( 58.20, 124) -- ( 76.20, 124);
\definecolor{drawColor}{RGB}{0,0,255}

\path[draw=drawColor,line width= 0.4pt,dash pattern=on 1pt off 3pt ,line join=round,line cap=round] ( 58.20, 104) -- ( 76.20, 104);
\definecolor{drawColor}{RGB}{0,255,0}

\path[draw=drawColor,line width= 0.4pt,dash pattern=on 1pt off 3pt ,line join=round,line cap=round] ( 58.20, 84) -- ( 76.20, 84);
\definecolor{drawColor}{RGB}{0,0,0}

\node[text=drawColor,anchor=base west,inner sep=0pt, outer sep=0pt, scale=  1] at ( 160.20,180) {Proposed};
\node[text=drawColor,anchor=base west,inner sep=0pt, outer sep=0pt, scale=  1] at ( 160.20,160) {Proposed};
\node[text=drawColor,anchor=base west,inner sep=0pt, outer sep=0pt, scale=  1] at ( 160.20,140) {Proposed};
\node[text=drawColor,anchor=base west,inner sep=0pt, outer sep=0pt, scale=  1] at ( 160.20,120) {Conventional};
\node[text=drawColor,anchor=base west,inner sep=0pt, outer sep=0pt, scale=  1] at ( 160.20,100) {Conventional};
\node[text=drawColor,anchor=base west,inner sep=0pt, outer sep=0pt, scale=  1] at ( 160.20,80) {Conventional};

\node[text=drawColor,anchor=base west,inner sep=0pt, outer sep=0pt, scale=  1] at ( 85.20,180) {$n=100$,};
\node[text=drawColor,anchor=base west,inner sep=0pt, outer sep=0pt, scale=  1] at ( 85.20,160) {$n=500$,};
\node[text=drawColor,anchor=base west,inner sep=0pt, outer sep=0pt, scale=  1] at ( 85.20,140) {$n=1000$,};

\node[text=drawColor,anchor=base west,inner sep=0pt, outer sep=0pt, scale=  1] at ( 85.20, 120) {$n=100$,};
\node[text=drawColor,anchor=base west,inner sep=0pt, outer sep=0pt, scale=  1] at ( 85.20, 100) {$n=500$,};
\node[text=drawColor,anchor=base west,inner sep=0pt, outer sep=0pt, scale=  1] at ( 85.20, 80) {$n=1000$,};
\end{scope}
\end{tikzpicture}

%% file: ec_rate1.tex
\begin{tikzpicture}[x=0.60pt,y=0.45pt]
\definecolor{fillColor}{RGB}{255,255,255}
\path[use as bounding box,fill=fillColor,fill opacity=0.00] (0,0) rectangle (361.35,361.35);
\begin{scope}
\path[clip] (  0.00,  0.00) rectangle (361.35,361.35);
\definecolor{drawColor}{RGB}{0,0,0}

\path[draw=drawColor,line width= 0.4pt,line join=round,line cap=round] ( 59.83, 61.20) -- (325.52, 61.20);

\path[draw=drawColor,line width= 0.4pt,line join=round,line cap=round] ( 59.83, 61.20) -- ( 59.83, 55.20);

\path[draw=drawColor,line width= 0.4pt,line join=round,line cap=round] (126.25, 61.20) -- (126.25, 55.20);

\path[draw=drawColor,line width= 0.4pt,line join=round,line cap=round] (192.68, 61.20) -- (192.68, 55.20);

\path[draw=drawColor,line width= 0.4pt,line join=round,line cap=round] (259.10, 61.20) -- (259.10, 55.20);

\path[draw=drawColor,line width= 0.4pt,line join=round,line cap=round] (325.52, 61.20) -- (325.52, 55.20);

\node[text=drawColor,anchor=base,inner sep=0pt, outer sep=0pt, scale=  1.00] at ( 59.83, 39.60) {2};

\node[text=drawColor,anchor=base,inner sep=0pt, outer sep=0pt, scale=  1.00] at (126.25, 39.60) {4};

\node[text=drawColor,anchor=base,inner sep=0pt, outer sep=0pt, scale=  1.00] at (192.68, 39.60) {6};

\node[text=drawColor,anchor=base,inner sep=0pt, outer sep=0pt, scale=  1.00] at (259.10, 39.60) {8};

\node[text=drawColor,anchor=base,inner sep=0pt, outer sep=0pt, scale=  1.00] at (325.52, 39.60) {10};

\path[draw=drawColor,line width= 0.4pt,line join=round,line cap=round] ( 49.20, 70.49) -- ( 49.20,302.86);

\path[draw=drawColor,line width= 0.4pt,line join=round,line cap=round] ( 49.20, 70.49) -- ( 43.20, 70.49);

\path[draw=drawColor,line width= 0.4pt,line join=round,line cap=round] ( 49.20,116.97) -- ( 43.20,116.97);

\path[draw=drawColor,line width= 0.4pt,line join=round,line cap=round] ( 49.20,163.44) -- ( 43.20,163.44);

\path[draw=drawColor,line width= 0.4pt,line join=round,line cap=round] ( 49.20,209.91) -- ( 43.20,209.91);

\path[draw=drawColor,line width= 0.4pt,line join=round,line cap=round] ( 49.20,256.38) -- ( 43.20,256.38);

\path[draw=drawColor,line width= 0.4pt,line join=round,line cap=round] ( 49.20,302.86) -- ( 43.20,302.86);

\node[text=drawColor,rotate= 90.00,anchor=base,inner sep=0pt, outer sep=0pt, scale=  1.00] at ( 34.80, 70.49) {0.0};

\node[text=drawColor,rotate= 90.00,anchor=base,inner sep=0pt, outer sep=0pt, scale=  1.00] at ( 34.80,116.97) {0.2};

\node[text=drawColor,rotate= 90.00,anchor=base,inner sep=0pt, outer sep=0pt, scale=  1.00] at ( 34.80,163.44) {0.4};

\node[text=drawColor,rotate= 90.00,anchor=base,inner sep=0pt, outer sep=0pt, scale=  1.00] at ( 34.80,209.91) {0.6};

\node[text=drawColor,rotate= 90.00,anchor=base,inner sep=0pt, outer sep=0pt, scale=  1.00] at ( 34.80,256.38) {0.8};

\node[text=drawColor,rotate= 90.00,anchor=base,inner sep=0pt, outer sep=0pt, scale=  1.00] at ( 34.80,302.86) {1.0};

\path[draw=drawColor,line width= 0.4pt,line join=round,line cap=round] ( 49.20, 61.20) --
	(336.15, 61.20) --
	(336.15,312.15) --
	( 49.20,312.15) --
	( 49.20, 61.20);
\end{scope}
\begin{scope}
\path[clip] (  0.00,  0.00) rectangle (361.35,361.35);
\definecolor{drawColor}{RGB}{0,0,0}

\node[text=drawColor,anchor=base,inner sep=0pt, outer sep=0pt, scale=  1.20] at (192.68,332.61) {\bfseries Local Confounding (Complete)};

\node[text=drawColor,anchor=base,inner sep=0pt, outer sep=0pt, scale=  1.00] at (192.68, 15.60) {The Number of  Variables p};

\node[text=drawColor,rotate= 90.00,anchor=base,inner sep=0pt, outer sep=0pt, scale=  1.00] at ( 10.80,186.67) {Correct Rate};
\end{scope}
\begin{scope}
\path[clip] ( 49.20, 61.20) rectangle (336.15,312.15);
\definecolor{drawColor}{RGB}{255,0,0}

\path[draw=drawColor,line width= 0.4pt,line join=round,line cap=round] ( 59.83,261.03) --
	(126.25,198.29) --
	(192.68,163.44) --
	(259.10, 84.44) --
	(325.52, 77.47);
\definecolor{drawColor}{RGB}{0,0,0}

\path[draw=drawColor,line width= 0.4pt,line join=round,line cap=round] ( 59.83,264.53) --
	( 62.86,259.28) --
	( 56.80,259.28) --
	( 59.83,264.53);

\path[draw=drawColor,line width= 0.4pt,line join=round,line cap=round] (126.25,201.79) --
	(129.28,196.54) --
	(123.22,196.54) --
	(126.25,201.79);

\path[draw=drawColor,line width= 0.4pt,line join=round,line cap=round] (192.68,166.94) --
	(195.71,161.69) --
	(189.64,161.69) --
	(192.68,166.94);

\path[draw=drawColor,line width= 0.4pt,line join=round,line cap=round] (259.10, 87.94) --
	(262.13, 82.69) --
	(256.07, 82.69) --
	(259.10, 87.94);

\path[draw=drawColor,line width= 0.4pt,line join=round,line cap=round] (325.52, 80.96) --
	(328.55, 75.72) --
	(322.49, 75.72) --
	(325.52, 80.96);
\definecolor{drawColor}{RGB}{0,0,255}

\path[draw=drawColor,line width= 0.4pt,line join=round,line cap=round] ( 59.83,270.32) --
	(126.25,254.06) --
	(192.68,198.29) --
	(259.10, 93.73) --
	(325.52, 82.11);
\definecolor{drawColor}{RGB}{0,0,0}

\path[draw=drawColor,line width= 0.4pt,line join=round,line cap=round] ( 56.65,270.32) -- ( 63.01,270.32);

\path[draw=drawColor,line width= 0.4pt,line join=round,line cap=round] ( 59.83,267.14) -- ( 59.83,273.51);

\path[draw=drawColor,line width= 0.4pt,line join=round,line cap=round] (123.07,254.06) -- (129.43,254.06);

\path[draw=drawColor,line width= 0.4pt,line join=round,line cap=round] (126.25,250.88) -- (126.25,257.24);

\path[draw=drawColor,line width= 0.4pt,line join=round,line cap=round] (189.49,198.29) -- (195.86,198.29);

\path[draw=drawColor,line width= 0.4pt,line join=round,line cap=round] (192.68,195.11) -- (192.68,201.48);

\path[draw=drawColor,line width= 0.4pt,line join=round,line cap=round] (255.92, 93.73) -- (262.28, 93.73);

\path[draw=drawColor,line width= 0.4pt,line join=round,line cap=round] (259.10, 90.55) -- (259.10, 96.91);

\path[draw=drawColor,line width= 0.4pt,line join=round,line cap=round] (322.34, 82.11) -- (328.70, 82.11);

\path[draw=drawColor,line width= 0.4pt,line join=round,line cap=round] (325.52, 78.93) -- (325.52, 85.29);
\definecolor{drawColor}{RGB}{0,255,0}

\path[draw=drawColor,line width= 0.4pt,line join=round,line cap=round] ( 59.83,281.94) --
	(126.25,240.12) --
	(192.68,193.65) --
	(259.10,107.67) --
	(325.52, 91.41);
\definecolor{drawColor}{RGB}{0,0,0}

\path[draw=drawColor,line width= 0.4pt,line join=round,line cap=round] ( 57.58,279.69) -- ( 62.08,284.19);

\path[draw=drawColor,line width= 0.4pt,line join=round,line cap=round] ( 57.58,284.19) -- ( 62.08,279.69);

\path[draw=drawColor,line width= 0.4pt,line join=round,line cap=round] (124.00,237.87) -- (128.50,242.37);

\path[draw=drawColor,line width= 0.4pt,line join=round,line cap=round] (124.00,242.37) -- (128.50,237.87);

\path[draw=drawColor,line width= 0.4pt,line join=round,line cap=round] (190.43,191.40) -- (194.93,195.90);

\path[draw=drawColor,line width= 0.4pt,line join=round,line cap=round] (190.43,195.90) -- (194.93,191.40);

\path[draw=drawColor,line width= 0.4pt,line join=round,line cap=round] (256.85,105.42) -- (261.35,109.92);

\path[draw=drawColor,line width= 0.4pt,line join=round,line cap=round] (256.85,109.92) -- (261.35,105.42);

\path[draw=drawColor,line width= 0.4pt,line join=round,line cap=round] (323.27, 89.16) -- (327.77, 93.66);

\path[draw=drawColor,line width= 0.4pt,line join=round,line cap=round] (323.27, 93.66) -- (327.77, 89.16);
\definecolor{drawColor}{RGB}{255,0,0}

\path[draw=drawColor,line width= 0.4pt,dash pattern=on 1pt off 3pt ,line join=round,line cap=round] ( 59.83,261.03) --
	(126.25,158.79) --
	(192.68, 89.08) --
	(259.10, 70.49) --
	(325.52, 70.49);
\definecolor{drawColor}{RGB}{0,0,0}

\path[draw=drawColor,line width= 0.4pt,line join=round,line cap=round] ( 56.65,261.03) --
	( 59.83,264.21) --
	( 63.01,261.03) --
	( 59.83,257.85) --
	( 56.65,261.03);

\path[draw=drawColor,line width= 0.4pt,line join=round,line cap=round] (123.07,158.79) --
	(126.25,161.97) --
	(129.43,158.79) --
	(126.25,155.61) --
	(123.07,158.79);

\path[draw=drawColor,line width= 0.4pt,line join=round,line cap=round] (189.49, 89.08) --
	(192.68, 92.27) --
	(195.86, 89.08) --
	(192.68, 85.90) --
	(189.49, 89.08);

\path[draw=drawColor,line width= 0.4pt,line join=round,line cap=round] (255.92, 70.49) --
	(259.10, 73.68) --
	(262.28, 70.49) --
	(259.10, 67.31) --
	(255.92, 70.49);

\path[draw=drawColor,line width= 0.4pt,line join=round,line cap=round] (322.34, 70.49) --
	(325.52, 73.68) --
	(328.70, 70.49) --
	(325.52, 67.31) --
	(322.34, 70.49);
\definecolor{drawColor}{RGB}{0,0,255}

\path[draw=drawColor,line width= 0.4pt,dash pattern=on 1pt off 3pt ,line join=round,line cap=round] ( 59.83,272.65) --
	(126.25,186.67) --
	(192.68,105.35) --
	(259.10, 70.49) --
	(325.52, 70.49);
\definecolor{drawColor}{RGB}{0,0,0}

\path[draw=drawColor,line width= 0.4pt,line join=round,line cap=round] ( 59.83,269.15) --
	( 62.86,274.40) --
	( 56.80,274.40) --
	( 59.83,269.15);

\path[draw=drawColor,line width= 0.4pt,line join=round,line cap=round] (126.25,183.18) --
	(129.28,188.42) --
	(123.22,188.42) --
	(126.25,183.18);

\path[draw=drawColor,line width= 0.4pt,line join=round,line cap=round] (192.68,101.85) --
	(195.71,107.10) --
	(189.64,107.10) --
	(192.68,101.85);

\path[draw=drawColor,line width= 0.4pt,line join=round,line cap=round] (259.10, 67.00) --
	(262.13, 72.24) --
	(256.07, 72.24) --
	(259.10, 67.00);

\path[draw=drawColor,line width= 0.4pt,line join=round,line cap=round] (325.52, 67.00) --
	(328.55, 72.24) --
	(322.49, 72.24) --
	(325.52, 67.00);
\definecolor{drawColor}{RGB}{0,255,0}

\path[draw=drawColor,line width= 0.4pt,dash pattern=on 1pt off 3pt ,line join=round,line cap=round] ( 59.83,277.30) --
	(126.25,200.62) --
	(192.68,121.61) --
	(259.10, 72.82) --
	(325.52, 70.49);
\definecolor{drawColor}{RGB}{0,0,0}

\path[draw=drawColor,line width= 0.4pt,line join=round,line cap=round] ( 57.58,275.05) rectangle ( 62.08,279.55);

\path[draw=drawColor,line width= 0.4pt,line join=round,line cap=round] ( 57.58,275.05) -- ( 62.08,279.55);

\path[draw=drawColor,line width= 0.4pt,line join=round,line cap=round] ( 57.58,279.55) -- ( 62.08,275.05);

\path[draw=drawColor,line width= 0.4pt,line join=round,line cap=round] (124.00,198.37) rectangle (128.50,202.87);

\path[draw=drawColor,line width= 0.4pt,line join=round,line cap=round] (124.00,198.37) -- (128.50,202.87);

\path[draw=drawColor,line width= 0.4pt,line join=round,line cap=round] (124.00,202.87) -- (128.50,198.37);

\path[draw=drawColor,line width= 0.4pt,line join=round,line cap=round] (190.43,119.36) rectangle (194.93,123.86);

\path[draw=drawColor,line width= 0.4pt,line join=round,line cap=round] (190.43,119.36) -- (194.93,123.86);

\path[draw=drawColor,line width= 0.4pt,line join=round,line cap=round] (190.43,123.86) -- (194.93,119.36);

\path[draw=drawColor,line width= 0.4pt,line join=round,line cap=round] (256.85, 70.57) rectangle (261.35, 75.07);

\path[draw=drawColor,line width= 0.4pt,line join=round,line cap=round] (256.85, 70.57) -- (261.35, 75.07);

\path[draw=drawColor,line width= 0.4pt,line join=round,line cap=round] (256.85, 75.07) -- (261.35, 70.57);

\path[draw=drawColor,line width= 0.4pt,line join=round,line cap=round] (323.27, 68.24) rectangle (327.77, 72.74);

\path[draw=drawColor,line width= 0.4pt,line join=round,line cap=round] (323.27, 68.24) -- (327.77, 72.74);

\path[draw=drawColor,line width= 0.4pt,line join=round,line cap=round] (323.27, 72.74) -- (327.77, 68.24);
\end{scope}
\end{tikzpicture}

%% file: ec_rate2.tex
\begin{tikzpicture}[x=0.60pt,y=0.45pt]
\definecolor{fillColor}{RGB}{255,255,255}
\path[use as bounding box,fill=fillColor,fill opacity=0.00] (0,0) rectangle (361.35,361.35);
\begin{scope}
\path[clip] (  0.00,  0.00) rectangle (361.35,361.35);
\definecolor{drawColor}{RGB}{0,0,0}

\path[draw=drawColor,line width= 0.4pt,line join=round,line cap=round] ( 59.83, 61.20) -- (325.52, 61.20);

\path[draw=drawColor,line width= 0.4pt,line join=round,line cap=round] ( 59.83, 61.20) -- ( 59.83, 55.20);

\path[draw=drawColor,line width= 0.4pt,line join=round,line cap=round] (126.25, 61.20) -- (126.25, 55.20);

\path[draw=drawColor,line width= 0.4pt,line join=round,line cap=round] (192.68, 61.20) -- (192.68, 55.20);

\path[draw=drawColor,line width= 0.4pt,line join=round,line cap=round] (259.10, 61.20) -- (259.10, 55.20);

\path[draw=drawColor,line width= 0.4pt,line join=round,line cap=round] (325.52, 61.20) -- (325.52, 55.20);

\node[text=drawColor,anchor=base,inner sep=0pt, outer sep=0pt, scale=  1.00] at ( 59.83, 39.60) {2};

\node[text=drawColor,anchor=base,inner sep=0pt, outer sep=0pt, scale=  1.00] at (126.25, 39.60) {4};

\node[text=drawColor,anchor=base,inner sep=0pt, outer sep=0pt, scale=  1.00] at (192.68, 39.60) {6};

\node[text=drawColor,anchor=base,inner sep=0pt, outer sep=0pt, scale=  1.00] at (259.10, 39.60) {8};

\node[text=drawColor,anchor=base,inner sep=0pt, outer sep=0pt, scale=  1.00] at (325.52, 39.60) {10};

\path[draw=drawColor,line width= 0.4pt,line join=round,line cap=round] ( 49.20, 70.49) -- ( 49.20,302.86);

\path[draw=drawColor,line width= 0.4pt,line join=round,line cap=round] ( 49.20, 70.49) -- ( 43.20, 70.49);

\path[draw=drawColor,line width= 0.4pt,line join=round,line cap=round] ( 49.20,116.97) -- ( 43.20,116.97);

\path[draw=drawColor,line width= 0.4pt,line join=round,line cap=round] ( 49.20,163.44) -- ( 43.20,163.44);

\path[draw=drawColor,line width= 0.4pt,line join=round,line cap=round] ( 49.20,209.91) -- ( 43.20,209.91);

\path[draw=drawColor,line width= 0.4pt,line join=round,line cap=round] ( 49.20,256.38) -- ( 43.20,256.38);

\path[draw=drawColor,line width= 0.4pt,line join=round,line cap=round] ( 49.20,302.86) -- ( 43.20,302.86);

\node[text=drawColor,rotate= 90.00,anchor=base,inner sep=0pt, outer sep=0pt, scale=  1.00] at ( 34.80, 70.49) {0.0};

\node[text=drawColor,rotate= 90.00,anchor=base,inner sep=0pt, outer sep=0pt, scale=  1.00] at ( 34.80,116.97) {0.2};

\node[text=drawColor,rotate= 90.00,anchor=base,inner sep=0pt, outer sep=0pt, scale=  1.00] at ( 34.80,163.44) {0.4};

\node[text=drawColor,rotate= 90.00,anchor=base,inner sep=0pt, outer sep=0pt, scale=  1.00] at ( 34.80,209.91) {0.6};

\node[text=drawColor,rotate= 90.00,anchor=base,inner sep=0pt, outer sep=0pt, scale=  1.00] at ( 34.80,256.38) {0.8};

\node[text=drawColor,rotate= 90.00,anchor=base,inner sep=0pt, outer sep=0pt, scale=  1.00] at ( 34.80,302.86) {1.0};

\path[draw=drawColor,line width= 0.4pt,line join=round,line cap=round] ( 49.20, 61.20) --
	(336.15, 61.20) --
	(336.15,312.15) --
	( 49.20,312.15) --
	( 49.20, 61.20);
\end{scope}
\begin{scope}
\path[clip] (  0.00,  0.00) rectangle (361.35,361.35);
\definecolor{drawColor}{RGB}{0,0,0}

\node[text=drawColor,anchor=base,inner sep=0pt, outer sep=0pt, scale=  1.20] at (192.68,332.61) {\bfseries Local Confounding (Pairwise)};

\node[text=drawColor,anchor=base,inner sep=0pt, outer sep=0pt, scale=  1.00] at (192.68, 15.60) {The Number of  Variables p};

\node[text=drawColor,rotate= 90.00,anchor=base,inner sep=0pt, outer sep=0pt, scale=  1.00] at ( 10.80,186.67) {Correct Rate};
\end{scope}
\begin{scope}
\path[clip] ( 49.20, 61.20) rectangle (336.15,312.15);
\definecolor{drawColor}{RGB}{255,0,0}

\path[draw=drawColor,line width= 0.4pt,line join=round,line cap=round] ( 59.83,261.03) --
	(126.25,268.78) --
	(192.68,267.54) --
	(259.10,223.02) --
	(325.52,214.97);
\definecolor{drawColor}{RGB}{0,0,0}

\path[draw=drawColor,line width= 0.4pt,line join=round,line cap=round] ( 59.83,264.53) --
	( 62.86,259.28) --
	( 56.80,259.28) --
	( 59.83,264.53);

\path[draw=drawColor,line width= 0.4pt,line join=round,line cap=round] (126.25,272.27) --
	(129.28,267.03) --
	(123.22,267.03) --
	(126.25,272.27);

\path[draw=drawColor,line width= 0.4pt,line join=round,line cap=round] (192.68,271.04) --
	(195.71,265.79) --
	(189.64,265.79) --
	(192.68,271.04);

\path[draw=drawColor,line width= 0.4pt,line join=round,line cap=round] (259.10,226.52) --
	(262.13,221.27) --
	(256.07,221.27) --
	(259.10,226.52);

\path[draw=drawColor,line width= 0.4pt,line join=round,line cap=round] (325.52,218.47) --
	(328.55,213.22) --
	(322.49,213.22) --
	(325.52,218.47);
\definecolor{drawColor}{RGB}{0,0,255}

\path[draw=drawColor,line width= 0.4pt,line join=round,line cap=round] ( 59.83,270.32) --
	(126.25,293.56) --
	(192.68,287.83) --
	(259.10,218.38) --
	(325.52,212.03);
\definecolor{drawColor}{RGB}{0,0,0}

\path[draw=drawColor,line width= 0.4pt,line join=round,line cap=round] ( 56.65,270.32) -- ( 63.01,270.32);

\path[draw=drawColor,line width= 0.4pt,line join=round,line cap=round] ( 59.83,267.14) -- ( 59.83,273.51);

\path[draw=drawColor,line width= 0.4pt,line join=round,line cap=round] (123.07,293.56) -- (129.43,293.56);

\path[draw=drawColor,line width= 0.4pt,line join=round,line cap=round] (126.25,290.38) -- (126.25,296.74);

\path[draw=drawColor,line width= 0.4pt,line join=round,line cap=round] (189.49,287.83) -- (195.86,287.83);

\path[draw=drawColor,line width= 0.4pt,line join=round,line cap=round] (192.68,284.65) -- (192.68,291.01);

\path[draw=drawColor,line width= 0.4pt,line join=round,line cap=round] (255.92,218.38) -- (262.28,218.38);

\path[draw=drawColor,line width= 0.4pt,line join=round,line cap=round] (259.10,215.19) -- (259.10,221.56);

\path[draw=drawColor,line width= 0.4pt,line join=round,line cap=round] (322.34,212.03) -- (328.70,212.03);

\path[draw=drawColor,line width= 0.4pt,line join=round,line cap=round] (325.52,208.85) -- (325.52,215.21);
\definecolor{drawColor}{RGB}{0,255,0}

\path[draw=drawColor,line width= 0.4pt,line join=round,line cap=round] ( 59.83,281.94) --
	(126.25,285.43) --
	(192.68,276.37) --
	(259.10,233.23) --
	(325.52,214.20);
\definecolor{drawColor}{RGB}{0,0,0}

\path[draw=drawColor,line width= 0.4pt,line join=round,line cap=round] ( 57.58,279.69) -- ( 62.08,284.19);

\path[draw=drawColor,line width= 0.4pt,line join=round,line cap=round] ( 57.58,284.19) -- ( 62.08,279.69);

\path[draw=drawColor,line width= 0.4pt,line join=round,line cap=round] (124.00,283.18) -- (128.50,287.68);

\path[draw=drawColor,line width= 0.4pt,line join=round,line cap=round] (124.00,287.68) -- (128.50,283.18);

\path[draw=drawColor,line width= 0.4pt,line join=round,line cap=round] (190.43,274.12) -- (194.93,278.62);

\path[draw=drawColor,line width= 0.4pt,line join=round,line cap=round] (190.43,278.62) -- (194.93,274.12);

\path[draw=drawColor,line width= 0.4pt,line join=round,line cap=round] (256.85,230.98) -- (261.35,235.48);

\path[draw=drawColor,line width= 0.4pt,line join=round,line cap=round] (256.85,235.48) -- (261.35,230.98);

\path[draw=drawColor,line width= 0.4pt,line join=round,line cap=round] (323.27,211.95) -- (327.77,216.45);

\path[draw=drawColor,line width= 0.4pt,line join=round,line cap=round] (323.27,216.45) -- (327.77,211.95);
\definecolor{drawColor}{RGB}{255,0,0}

\path[draw=drawColor,line width= 0.4pt,dash pattern=on 1pt off 3pt ,line join=round,line cap=round] ( 59.83,261.03) --
	(126.25,260.64) --
	(192.68,254.06) --
	(259.10,203.27) --
	(325.52,201.39);
\definecolor{drawColor}{RGB}{0,0,0}

\path[draw=drawColor,line width= 0.4pt,line join=round,line cap=round] ( 56.65,261.03) --
	( 59.83,264.21) --
	( 63.01,261.03) --
	( 59.83,257.85) --
	( 56.65,261.03);

\path[draw=drawColor,line width= 0.4pt,line join=round,line cap=round] (123.07,260.64) --
	(126.25,263.83) --
	(129.43,260.64) --
	(126.25,257.46) --
	(123.07,260.64);

\path[draw=drawColor,line width= 0.4pt,line join=round,line cap=round] (189.49,254.06) --
	(192.68,257.24) --
	(195.86,254.06) --
	(192.68,250.88) --
	(189.49,254.06);

\path[draw=drawColor,line width= 0.4pt,line join=round,line cap=round] (255.92,203.27) --
	(259.10,206.45) --
	(262.28,203.27) --
	(259.10,200.09) --
	(255.92,203.27);

\path[draw=drawColor,line width= 0.4pt,line join=round,line cap=round] (322.34,201.39) --
	(325.52,204.57) --
	(328.70,201.39) --
	(325.52,198.21) --
	(322.34,201.39);
\definecolor{drawColor}{RGB}{0,0,255}

\path[draw=drawColor,line width= 0.4pt,dash pattern=on 1pt off 3pt ,line join=round,line cap=round] ( 59.83,272.65) --
	(126.25,278.84) --
	(192.68,267.69) --
	(259.10,207.67) --
	(325.52,210.94);
\definecolor{drawColor}{RGB}{0,0,0}

\path[draw=drawColor,line width= 0.4pt,line join=round,line cap=round] ( 59.83,269.15) --
	( 62.86,274.40) --
	( 56.80,274.40) --
	( 59.83,269.15);

\path[draw=drawColor,line width= 0.4pt,line join=round,line cap=round] (126.25,275.35) --
	(129.28,280.59) --
	(123.22,280.59) --
	(126.25,275.35);

\path[draw=drawColor,line width= 0.4pt,line join=round,line cap=round] (192.68,264.19) --
	(195.71,269.44) --
	(189.64,269.44) --
	(192.68,264.19);

\path[draw=drawColor,line width= 0.4pt,line join=round,line cap=round] (259.10,204.17) --
	(262.13,209.42) --
	(256.07,209.42) --
	(259.10,204.17);

\path[draw=drawColor,line width= 0.4pt,line join=round,line cap=round] (325.52,207.44) --
	(328.55,212.69) --
	(322.49,212.69) --
	(325.52,207.44);
\definecolor{drawColor}{RGB}{0,255,0}

\path[draw=drawColor,line width= 0.4pt,dash pattern=on 1pt off 3pt ,line join=round,line cap=round] ( 59.83,277.30) --
	(126.25,281.17) --
	(192.68,272.96) --
	(259.10,211.07) --
	(325.52,211.92);
\definecolor{drawColor}{RGB}{0,0,0}

\path[draw=drawColor,line width= 0.4pt,line join=round,line cap=round] ( 57.58,275.05) rectangle ( 62.08,279.55);

\path[draw=drawColor,line width= 0.4pt,line join=round,line cap=round] ( 57.58,275.05) -- ( 62.08,279.55);

\path[draw=drawColor,line width= 0.4pt,line join=round,line cap=round] ( 57.58,279.55) -- ( 62.08,275.05);

\path[draw=drawColor,line width= 0.4pt,line join=round,line cap=round] (124.00,278.92) rectangle (128.50,283.42);

\path[draw=drawColor,line width= 0.4pt,line join=round,line cap=round] (124.00,278.92) -- (128.50,283.42);

\path[draw=drawColor,line width= 0.4pt,line join=round,line cap=round] (124.00,283.42) -- (128.50,278.92);

\path[draw=drawColor,line width= 0.4pt,line join=round,line cap=round] (190.43,270.71) rectangle (194.93,275.21);

\path[draw=drawColor,line width= 0.4pt,line join=round,line cap=round] (190.43,270.71) -- (194.93,275.21);

\path[draw=drawColor,line width= 0.4pt,line join=round,line cap=round] (190.43,275.21) -- (194.93,270.71);

\path[draw=drawColor,line width= 0.4pt,line join=round,line cap=round] (256.85,208.82) rectangle (261.35,213.32);

\path[draw=drawColor,line width= 0.4pt,line join=round,line cap=round] (256.85,208.82) -- (261.35,213.32);

\path[draw=drawColor,line width= 0.4pt,line join=round,line cap=round] (256.85,213.32) -- (261.35,208.82);

\path[draw=drawColor,line width= 0.4pt,line join=round,line cap=round] (323.27,209.67) rectangle (327.77,214.17);

\path[draw=drawColor,line width= 0.4pt,line join=round,line cap=round] (323.27,209.67) -- (327.77,214.17);

\path[draw=drawColor,line width= 0.4pt,line join=round,line cap=round] (323.27,214.17) -- (327.77,209.67);
\definecolor{drawColor}{RGB}{255,0,0}

\path[draw=drawColor,line width= 0.4pt,line join=round,line cap=round] ( 58.20,184) -- ( 76.20,184);
\definecolor{drawColor}{RGB}{0,0,255}

\path[draw=drawColor,line width= 0.4pt,line join=round,line cap=round] ( 58.20,164) -- ( 76.20,164);
\definecolor{drawColor}{RGB}{0,255,0}

\path[draw=drawColor,line width= 0.4pt,line join=round,line cap=round] ( 58.20,144) -- ( 76.20,144);
\definecolor{drawColor}{RGB}{255,0,0}

\path[draw=drawColor,line width= 0.4pt,dash pattern=on 1pt off 3pt ,line join=round,line cap=round] ( 58.20, 124) -- ( 76.20, 124);
\definecolor{drawColor}{RGB}{0,0,255}

\path[draw=drawColor,line width= 0.4pt,dash pattern=on 1pt off 3pt ,line join=round,line cap=round] ( 58.20, 104) -- ( 76.20, 104);
\definecolor{drawColor}{RGB}{0,255,0}

\path[draw=drawColor,line width= 0.4pt,dash pattern=on 1pt off 3pt ,line join=round,line cap=round] ( 58.20, 84) -- ( 76.20, 84);
\definecolor{drawColor}{RGB}{0,0,0}

\node[text=drawColor,anchor=base west,inner sep=0pt, outer sep=0pt, scale=  1.00] at ( 160.20,180) {Proposed};
\node[text=drawColor,anchor=base west,inner sep=0pt, outer sep=0pt, scale=  1.00] at ( 160.20,160) {Proposed};
\node[text=drawColor,anchor=base west,inner sep=0pt, outer sep=0pt, scale=  1.00] at ( 160.20,140) {Proposed};
\node[text=drawColor,anchor=base west,inner sep=0pt, outer sep=0pt, scale=  1.00] at ( 160.20,120) {Conventional};
\node[text=drawColor,anchor=base west,inner sep=0pt, outer sep=0pt, scale=  1.00] at ( 160.20,100) {Conventional};
\node[text=drawColor,anchor=base west,inner sep=0pt, outer sep=0pt, scale=  1.00] at ( 160.20,80) {Conventional};

\node[text=drawColor,anchor=base west,inner sep=0pt, outer sep=0pt, scale=  1.00] at ( 85.20,180) {$n=100$,};
\node[text=drawColor,anchor=base west,inner sep=0pt, outer sep=0pt, scale=  1.00] at ( 85.20,160) {$n=500$,};
\node[text=drawColor,anchor=base west,inner sep=0pt, outer sep=0pt, scale=  1.00] at ( 85.20,140) {$n=1000$,};
\node[text=drawColor,anchor=base west,inner sep=0pt, outer sep=0pt, scale=  1.00] at ( 85.20, 120) {$n=100$,};
\node[text=drawColor,anchor=base west,inner sep=0pt, outer sep=0pt, scale=  1.00] at ( 85.20, 100) {$n=500$,};
\node[text=drawColor,anchor=base west,inner sep=0pt, outer sep=0pt, scale=  1.00] at ( 85.20, 80) {$n=1000$,};
\end{scope}
\end{tikzpicture}

%% file: ec2_rate1.tex
\begin{tikzpicture}[x=0.60pt,y=0.45pt]
\definecolor{fillColor}{RGB}{255,255,255}
\path[use as bounding box,fill=fillColor,fill opacity=0.00] (0,0) rectangle (361.35,361.35);
\begin{scope}
\path[clip] (  0.00,  0.00) rectangle (361.35,361.35);
\definecolor{drawColor}{RGB}{0,0,0}

\path[draw=drawColor,line width= 0.4pt,line join=round,line cap=round] ( 59.83, 61.20) -- (325.52, 61.20);

\path[draw=drawColor,line width= 0.4pt,line join=round,line cap=round] ( 59.83, 61.20) -- ( 59.83, 55.20);

\path[draw=drawColor,line width= 0.4pt,line join=round,line cap=round] (126.25, 61.20) -- (126.25, 55.20);

\path[draw=drawColor,line width= 0.4pt,line join=round,line cap=round] (192.68, 61.20) -- (192.68, 55.20);

\path[draw=drawColor,line width= 0.4pt,line join=round,line cap=round] (259.10, 61.20) -- (259.10, 55.20);

\path[draw=drawColor,line width= 0.4pt,line join=round,line cap=round] (325.52, 61.20) -- (325.52, 55.20);

\node[text=drawColor,anchor=base,inner sep=0pt, outer sep=0pt, scale=  1.00] at ( 59.83, 39.60) {4};

\node[text=drawColor,anchor=base,inner sep=0pt, outer sep=0pt, scale=  1.00] at (126.25, 39.60) {5};

\node[text=drawColor,anchor=base,inner sep=0pt, outer sep=0pt, scale=  1.00] at (192.68, 39.60) {6};

\node[text=drawColor,anchor=base,inner sep=0pt, outer sep=0pt, scale=  1.00] at (259.10, 39.60) {7};

\node[text=drawColor,anchor=base,inner sep=0pt, outer sep=0pt, scale=  1.00] at (325.52, 39.60) {8};

\path[draw=drawColor,line width= 0.4pt,line join=round,line cap=round] ( 49.20, 70.49) -- ( 49.20,302.86);

\path[draw=drawColor,line width= 0.4pt,line join=round,line cap=round] ( 49.20, 70.49) -- ( 43.20, 70.49);

\path[draw=drawColor,line width= 0.4pt,line join=round,line cap=round] ( 49.20,116.97) -- ( 43.20,116.97);

\path[draw=drawColor,line width= 0.4pt,line join=round,line cap=round] ( 49.20,163.44) -- ( 43.20,163.44);

\path[draw=drawColor,line width= 0.4pt,line join=round,line cap=round] ( 49.20,209.91) -- ( 43.20,209.91);

\path[draw=drawColor,line width= 0.4pt,line join=round,line cap=round] ( 49.20,256.38) -- ( 43.20,256.38);

\path[draw=drawColor,line width= 0.4pt,line join=round,line cap=round] ( 49.20,302.86) -- ( 43.20,302.86);

\node[text=drawColor,rotate= 90.00,anchor=base,inner sep=0pt, outer sep=0pt, scale=  1.00] at ( 34.80, 70.49) {0.0};

\node[text=drawColor,rotate= 90.00,anchor=base,inner sep=0pt, outer sep=0pt, scale=  1.00] at ( 34.80,116.97) {0.2};

\node[text=drawColor,rotate= 90.00,anchor=base,inner sep=0pt, outer sep=0pt, scale=  1.00] at ( 34.80,163.44) {0.4};

\node[text=drawColor,rotate= 90.00,anchor=base,inner sep=0pt, outer sep=0pt, scale=  1.00] at ( 34.80,209.91) {0.6};

\node[text=drawColor,rotate= 90.00,anchor=base,inner sep=0pt, outer sep=0pt, scale=  1.00] at ( 34.80,256.38) {0.8};

\node[text=drawColor,rotate= 90.00,anchor=base,inner sep=0pt, outer sep=0pt, scale=  1.00] at ( 34.80,302.86) {1.0};

\path[draw=drawColor,line width= 0.4pt,line join=round,line cap=round] ( 49.20, 61.20) --
	(336.15, 61.20) --
	(336.15,312.15) --
	( 49.20,312.15) --
	( 49.20, 61.20);
\end{scope}
\begin{scope}
\path[clip] (  0.00,  0.00) rectangle (361.35,361.35);
\definecolor{drawColor}{RGB}{0,0,0}

\node[text=drawColor,anchor=base,inner sep=0pt, outer sep=0pt, scale=  1.20] at (192.68,332.61) {\bfseries Global Confounding (Complete)};

\node[text=drawColor,anchor=base,inner sep=0pt, outer sep=0pt, scale=  1.00] at (192.68, 15.60) {The Number of  Variables p};

\node[text=drawColor,rotate= 90.00,anchor=base,inner sep=0pt, outer sep=0pt, scale=  1.00] at ( 10.80,186.67) {Correct Rate};
\end{scope}
\begin{scope}
\path[clip] ( 49.20, 61.20) rectangle (336.15,312.15);
\definecolor{drawColor}{RGB}{255,0,0}

\path[draw=drawColor,line width= 0.4pt,line join=round,line cap=round] ( 59.83,212.23) --
	(192.68,172.73) --
	(325.52, 84.44);
\definecolor{drawColor}{RGB}{0,0,0}

\path[draw=drawColor,line width= 0.4pt,line join=round,line cap=round] ( 59.83,215.73) --
	( 62.86,210.49) --
	( 56.80,210.49) --
	( 59.83,215.73);

\path[draw=drawColor,line width= 0.4pt,line join=round,line cap=round] (192.68,176.23) --
	(195.71,170.98) --
	(189.64,170.98) --
	(192.68,176.23);

\path[draw=drawColor,line width= 0.4pt,line join=round,line cap=round] (325.52, 87.94) --
	(328.55, 82.69) --
	(322.49, 82.69) --
	(325.52, 87.94);
\definecolor{drawColor}{RGB}{0,0,255}

\path[draw=drawColor,line width= 0.4pt,line join=round,line cap=round] ( 59.83,230.82) --
	(192.68,265.68) --
	(325.52,100.70);
\definecolor{drawColor}{RGB}{0,0,0}

\path[draw=drawColor,line width= 0.4pt,line join=round,line cap=round] ( 56.65,230.82) -- ( 63.01,230.82);

\path[draw=drawColor,line width= 0.4pt,line join=round,line cap=round] ( 59.83,227.64) -- ( 59.83,234.01);

\path[draw=drawColor,line width= 0.4pt,line join=round,line cap=round] (189.49,265.68) -- (195.86,265.68);

\path[draw=drawColor,line width= 0.4pt,line join=round,line cap=round] (192.68,262.50) -- (192.68,268.86);

\path[draw=drawColor,line width= 0.4pt,line join=round,line cap=round] (322.34,100.70) -- (328.70,100.70);

\path[draw=drawColor,line width= 0.4pt,line join=round,line cap=round] (325.52, 97.52) -- (325.52,103.88);
\definecolor{drawColor}{RGB}{0,255,0}

\path[draw=drawColor,line width= 0.4pt,line join=round,line cap=round] ( 59.83,249.41) --
	(192.68,265.68) --
	(325.52, 98.38);
\definecolor{drawColor}{RGB}{0,0,0}

\path[draw=drawColor,line width= 0.4pt,line join=round,line cap=round] ( 57.58,247.16) -- ( 62.08,251.66);

\path[draw=drawColor,line width= 0.4pt,line join=round,line cap=round] ( 57.58,251.66) -- ( 62.08,247.16);

\path[draw=drawColor,line width= 0.4pt,line join=round,line cap=round] (190.43,263.43) -- (194.93,267.93);

\path[draw=drawColor,line width= 0.4pt,line join=round,line cap=round] (190.43,267.93) -- (194.93,263.43);

\path[draw=drawColor,line width= 0.4pt,line join=round,line cap=round] (323.27, 96.13) -- (327.77,100.63);

\path[draw=drawColor,line width= 0.4pt,line join=round,line cap=round] (323.27,100.63) -- (327.77, 96.13);
\definecolor{drawColor}{RGB}{255,0,0}

\path[draw=drawColor,line width= 0.4pt,dash pattern=on 1pt off 3pt ,line join=round,line cap=round] ( 59.83,156.47) --
	(192.68, 96.05) --
	(325.52, 70.49);
\definecolor{drawColor}{RGB}{0,0,0}

\path[draw=drawColor,line width= 0.4pt,line join=round,line cap=round] ( 56.65,156.47) --
	( 59.83,159.65) --
	( 63.01,156.47) --
	( 59.83,153.29) --
	( 56.65,156.47);

\path[draw=drawColor,line width= 0.4pt,line join=round,line cap=round] (189.49, 96.05) --
	(192.68, 99.24) --
	(195.86, 96.05) --
	(192.68, 92.87) --
	(189.49, 96.05);

\path[draw=drawColor,line width= 0.4pt,line join=round,line cap=round] (322.34, 70.49) --
	(325.52, 73.68) --
	(328.70, 70.49) --
	(325.52, 67.31) --
	(322.34, 70.49);
\definecolor{drawColor}{RGB}{0,0,255}

\path[draw=drawColor,line width= 0.4pt,dash pattern=on 1pt off 3pt ,line join=round,line cap=round] ( 59.83,207.59) --
	(192.68,126.26) --
	(325.52, 70.49);
\definecolor{drawColor}{RGB}{0,0,0}

\path[draw=drawColor,line width= 0.4pt,line join=round,line cap=round] ( 59.83,204.09) --
	( 62.86,209.34) --
	( 56.80,209.34) --
	( 59.83,204.09);

\path[draw=drawColor,line width= 0.4pt,line join=round,line cap=round] (192.68,122.76) --
	(195.71,128.01) --
	(189.64,128.01) --
	(192.68,122.76);

\path[draw=drawColor,line width= 0.4pt,line join=round,line cap=round] (325.52, 67.00) --
	(328.55, 72.24) --
	(322.49, 72.24) --
	(325.52, 67.00);
\definecolor{drawColor}{RGB}{0,255,0}

\path[draw=drawColor,line width= 0.4pt,dash pattern=on 1pt off 3pt ,line join=round,line cap=round] ( 59.83,221.53) --
	(192.68,137.88) --
	(325.52, 70.49);
\definecolor{drawColor}{RGB}{0,0,0}

\path[draw=drawColor,line width= 0.4pt,line join=round,line cap=round] ( 57.58,219.28) rectangle ( 62.08,223.78);

\path[draw=drawColor,line width= 0.4pt,line join=round,line cap=round] ( 57.58,219.28) -- ( 62.08,223.78);

\path[draw=drawColor,line width= 0.4pt,line join=round,line cap=round] ( 57.58,223.78) -- ( 62.08,219.28);

\path[draw=drawColor,line width= 0.4pt,line join=round,line cap=round] (190.43,135.63) rectangle (194.93,140.13);

\path[draw=drawColor,line width= 0.4pt,line join=round,line cap=round] (190.43,135.63) -- (194.93,140.13);

\path[draw=drawColor,line width= 0.4pt,line join=round,line cap=round] (190.43,140.13) -- (194.93,135.63);

\path[draw=drawColor,line width= 0.4pt,line join=round,line cap=round] (323.27, 68.24) rectangle (327.77, 72.74);

\path[draw=drawColor,line width= 0.4pt,line join=round,line cap=round] (323.27, 68.24) -- (327.77, 72.74);

\path[draw=drawColor,line width= 0.4pt,line join=round,line cap=round] (323.27, 72.74) -- (327.77, 68.24);
\end{scope}
\end{tikzpicture}

%% file: ec2_rate2.tex
\begin{tikzpicture}[x=0.60pt,y=0.45pt]
\definecolor{fillColor}{RGB}{255,255,255}
\path[use as bounding box,fill=fillColor,fill opacity=0.00] (0,0) rectangle (361.35,361.35);
\begin{scope}
\path[clip] (  0.00,  0.00) rectangle (361.35,361.35);
\definecolor{drawColor}{RGB}{0,0,0}

\path[draw=drawColor,line width= 0.4pt,line join=round,line cap=round] ( 59.83, 61.20) -- (325.52, 61.20);

\path[draw=drawColor,line width= 0.4pt,line join=round,line cap=round] ( 59.83, 61.20) -- ( 59.83, 55.20);

\path[draw=drawColor,line width= 0.4pt,line join=round,line cap=round] (126.25, 61.20) -- (126.25, 55.20);

\path[draw=drawColor,line width= 0.4pt,line join=round,line cap=round] (192.68, 61.20) -- (192.68, 55.20);

\path[draw=drawColor,line width= 0.4pt,line join=round,line cap=round] (259.10, 61.20) -- (259.10, 55.20);

\path[draw=drawColor,line width= 0.4pt,line join=round,line cap=round] (325.52, 61.20) -- (325.52, 55.20);

\node[text=drawColor,anchor=base,inner sep=0pt, outer sep=0pt, scale=  1.00] at ( 59.83, 39.60) {4};

\node[text=drawColor,anchor=base,inner sep=0pt, outer sep=0pt, scale=  1.00] at (126.25, 39.60) {5};

\node[text=drawColor,anchor=base,inner sep=0pt, outer sep=0pt, scale=  1.00] at (192.68, 39.60) {6};

\node[text=drawColor,anchor=base,inner sep=0pt, outer sep=0pt, scale=  1.00] at (259.10, 39.60) {7};

\node[text=drawColor,anchor=base,inner sep=0pt, outer sep=0pt, scale=  1.00] at (325.52, 39.60) {8};

\path[draw=drawColor,line width= 0.4pt,line join=round,line cap=round] ( 49.20, 70.49) -- ( 49.20,302.86);

\path[draw=drawColor,line width= 0.4pt,line join=round,line cap=round] ( 49.20, 70.49) -- ( 43.20, 70.49);

\path[draw=drawColor,line width= 0.4pt,line join=round,line cap=round] ( 49.20,116.97) -- ( 43.20,116.97);

\path[draw=drawColor,line width= 0.4pt,line join=round,line cap=round] ( 49.20,163.44) -- ( 43.20,163.44);

\path[draw=drawColor,line width= 0.4pt,line join=round,line cap=round] ( 49.20,209.91) -- ( 43.20,209.91);

\path[draw=drawColor,line width= 0.4pt,line join=round,line cap=round] ( 49.20,256.38) -- ( 43.20,256.38);

\path[draw=drawColor,line width= 0.4pt,line join=round,line cap=round] ( 49.20,302.86) -- ( 43.20,302.86);

\node[text=drawColor,rotate= 90.00,anchor=base,inner sep=0pt, outer sep=0pt, scale=  1.00] at ( 34.80, 70.49) {0.0};

\node[text=drawColor,rotate= 90.00,anchor=base,inner sep=0pt, outer sep=0pt, scale=  1.00] at ( 34.80,116.97) {0.2};

\node[text=drawColor,rotate= 90.00,anchor=base,inner sep=0pt, outer sep=0pt, scale=  1.00] at ( 34.80,163.44) {0.4};

\node[text=drawColor,rotate= 90.00,anchor=base,inner sep=0pt, outer sep=0pt, scale=  1.00] at ( 34.80,209.91) {0.6};

\node[text=drawColor,rotate= 90.00,anchor=base,inner sep=0pt, outer sep=0pt, scale=  1.00] at ( 34.80,256.38) {0.8};

\node[text=drawColor,rotate= 90.00,anchor=base,inner sep=0pt, outer sep=0pt, scale=  1.00] at ( 34.80,302.86) {1.0};

\path[draw=drawColor,line width= 0.4pt,line join=round,line cap=round] ( 49.20, 61.20) --
	(336.15, 61.20) --
	(336.15,312.15) --
	( 49.20,312.15) --
	( 49.20, 61.20);
\end{scope}
\begin{scope}
\path[clip] (  0.00,  0.00) rectangle (361.35,361.35);
\definecolor{drawColor}{RGB}{0,0,0}

\node[text=drawColor,anchor=base,inner sep=0pt, outer sep=0pt, scale=  1.20] at (192.68,332.61) {\bfseries Global Confounding (Pairwise)};

\node[text=drawColor,anchor=base,inner sep=0pt, outer sep=0pt, scale=  1.00] at (192.68, 15.60) {The Number of  Variables p};

\node[text=drawColor,rotate= 90.00,anchor=base,inner sep=0pt, outer sep=0pt, scale=  1.00] at ( 10.80,186.67) {Correct Rate};
\end{scope}
\begin{scope}
\path[clip] ( 49.20, 61.20) rectangle (336.15,312.15);
\definecolor{drawColor}{RGB}{255,0,0}

\path[draw=drawColor,line width= 0.4pt,line join=round,line cap=round] ( 59.83,276.91) --
	(192.68,254.83) --
	(325.52,213.40);
\definecolor{drawColor}{RGB}{0,0,0}

\path[draw=drawColor,line width= 0.4pt,line join=round,line cap=round] ( 59.83,280.41) --
	( 62.86,275.16) --
	( 56.80,275.16) --
	( 59.83,280.41);

\path[draw=drawColor,line width= 0.4pt,line join=round,line cap=round] (192.68,258.33) --
	(195.71,253.08) --
	(189.64,253.08) --
	(192.68,258.33);

\path[draw=drawColor,line width= 0.4pt,line join=round,line cap=round] (325.52,216.90) --
	(328.55,211.65) --
	(322.49,211.65) --
	(325.52,216.90);
\definecolor{drawColor}{RGB}{0,0,255}

\path[draw=drawColor,line width= 0.4pt,line join=round,line cap=round] ( 59.83,285.43) --
	(192.68,300.22) --
	(325.52,223.85);
\definecolor{drawColor}{RGB}{0,0,0}

\path[draw=drawColor,line width= 0.4pt,line join=round,line cap=round] ( 56.65,285.43) -- ( 63.01,285.43);

\path[draw=drawColor,line width= 0.4pt,line join=round,line cap=round] ( 59.83,282.25) -- ( 59.83,288.61);

\path[draw=drawColor,line width= 0.4pt,line join=round,line cap=round] (189.49,300.22) -- (195.86,300.22);

\path[draw=drawColor,line width= 0.4pt,line join=round,line cap=round] (192.68,297.04) -- (192.68,303.40);

\path[draw=drawColor,line width= 0.4pt,line join=round,line cap=round] (322.34,223.85) -- (328.70,223.85);

\path[draw=drawColor,line width= 0.4pt,line join=round,line cap=round] (325.52,220.67) -- (325.52,227.03);
\definecolor{drawColor}{RGB}{0,255,0}

\path[draw=drawColor,line width= 0.4pt,line join=round,line cap=round] ( 59.83,289.69) --
	(192.68,300.22) --
	(325.52,224.85);
\definecolor{drawColor}{RGB}{0,0,0}

\path[draw=drawColor,line width= 0.4pt,line join=round,line cap=round] ( 57.58,287.44) -- ( 62.08,291.94);

\path[draw=drawColor,line width= 0.4pt,line join=round,line cap=round] ( 57.58,291.94) -- ( 62.08,287.44);

\path[draw=drawColor,line width= 0.4pt,line join=round,line cap=round] (190.43,297.97) -- (194.93,302.47);

\path[draw=drawColor,line width= 0.4pt,line join=round,line cap=round] (190.43,302.47) -- (194.93,297.97);

\path[draw=drawColor,line width= 0.4pt,line join=round,line cap=round] (323.27,222.60) -- (327.77,227.10);

\path[draw=drawColor,line width= 0.4pt,line join=round,line cap=round] (323.27,227.10) -- (327.77,222.60);
\definecolor{drawColor}{RGB}{255,0,0}

\path[draw=drawColor,line width= 0.4pt,dash pattern=on 1pt off 3pt ,line join=round,line cap=round] ( 59.83,262.97) --
	(192.68,248.48) --
	(325.52,200.70);
\definecolor{drawColor}{RGB}{0,0,0}

\path[draw=drawColor,line width= 0.4pt,line join=round,line cap=round] ( 56.65,262.97) --
	( 59.83,266.15) --
	( 63.01,262.97) --
	( 59.83,259.78) --
	( 56.65,262.97);

\path[draw=drawColor,line width= 0.4pt,line join=round,line cap=round] (189.49,248.48) --
	(192.68,251.67) --
	(195.86,248.48) --
	(192.68,245.30) --
	(189.49,248.48);

\path[draw=drawColor,line width= 0.4pt,line join=round,line cap=round] (322.34,200.70) --
	(325.52,203.88) --
	(328.70,200.70) --
	(325.52,197.52) --
	(322.34,200.70);
\definecolor{drawColor}{RGB}{0,0,255}

\path[draw=drawColor,line width= 0.4pt,dash pattern=on 1pt off 3pt ,line join=round,line cap=round] ( 59.83,284.65) --
	(192.68,274.04) --
	(325.52,207.34);
\definecolor{drawColor}{RGB}{0,0,0}

\path[draw=drawColor,line width= 0.4pt,line join=round,line cap=round] ( 59.83,281.15) --
	( 62.86,286.40) --
	( 56.80,286.40) --
	( 59.83,281.15);

\path[draw=drawColor,line width= 0.4pt,line join=round,line cap=round] (192.68,270.54) --
	(195.71,275.79) --
	(189.64,275.79) --
	(192.68,270.54);

\path[draw=drawColor,line width= 0.4pt,line join=round,line cap=round] (325.52,203.84) --
	(328.55,209.09) --
	(322.49,209.09) --
	(325.52,203.84);
\definecolor{drawColor}{RGB}{0,255,0}

\path[draw=drawColor,line width= 0.4pt,dash pattern=on 1pt off 3pt ,line join=round,line cap=round] ( 59.83,288.14) --
	(192.68,276.21) --
	(325.52,215.06);
\definecolor{drawColor}{RGB}{0,0,0}

\path[draw=drawColor,line width= 0.4pt,line join=round,line cap=round] ( 57.58,285.89) rectangle ( 62.08,290.39);

\path[draw=drawColor,line width= 0.4pt,line join=round,line cap=round] ( 57.58,285.89) -- ( 62.08,290.39);

\path[draw=drawColor,line width= 0.4pt,line join=round,line cap=round] ( 57.58,290.39) -- ( 62.08,285.89);

\path[draw=drawColor,line width= 0.4pt,line join=round,line cap=round] (190.43,273.96) rectangle (194.93,278.46);

\path[draw=drawColor,line width= 0.4pt,line join=round,line cap=round] (190.43,273.96) -- (194.93,278.46);

\path[draw=drawColor,line width= 0.4pt,line join=round,line cap=round] (190.43,278.46) -- (194.93,273.96);

\path[draw=drawColor,line width= 0.4pt,line join=round,line cap=round] (323.27,212.81) rectangle (327.77,217.31);

\path[draw=drawColor,line width= 0.4pt,line join=round,line cap=round] (323.27,212.81) -- (327.77,217.31);

\path[draw=drawColor,line width= 0.4pt,line join=round,line cap=round] (323.27,217.31) -- (327.77,212.81);
\definecolor{drawColor}{RGB}{255,0,0}

\path[draw=drawColor,line width= 0.4pt,line join=round,line cap=round] ( 58.20,184) -- ( 76.20,184);
\definecolor{drawColor}{RGB}{0,0,255}

\path[draw=drawColor,line width= 0.4pt,line join=round,line cap=round] ( 58.20,164) -- ( 76.20,164);
\definecolor{drawColor}{RGB}{0,255,0}

\path[draw=drawColor,line width= 0.4pt,line join=round,line cap=round] ( 58.20,144) -- ( 76.20,144);
\definecolor{drawColor}{RGB}{255,0,0}

\path[draw=drawColor,line width= 0.4pt,dash pattern=on 1pt off 3pt ,line join=round,line cap=round] ( 58.20, 124) -- ( 76.20, 124);
\definecolor{drawColor}{RGB}{0,0,255}

\path[draw=drawColor,line width= 0.4pt,dash pattern=on 1pt off 3pt ,line join=round,line cap=round] ( 58.20, 104) -- ( 76.20, 104);
\definecolor{drawColor}{RGB}{0,255,0}

\path[draw=drawColor,line width= 0.4pt,dash pattern=on 1pt off 3pt ,line join=round,line cap=round] ( 58.20, 84) -- ( 76.20, 84);
\definecolor{drawColor}{RGB}{0,0,0}

\node[text=drawColor,anchor=base west,inner sep=0pt, outer sep=0pt, scale=  1.00] at ( 160.20,180) {Proposed};
\node[text=drawColor,anchor=base west,inner sep=0pt, outer sep=0pt, scale=  1.00] at ( 160.20,160) {Proposed};
\node[text=drawColor,anchor=base west,inner sep=0pt, outer sep=0pt, scale=  1.00] at ( 160.20,140) {Proposed};
\node[text=drawColor,anchor=base west,inner sep=0pt, outer sep=0pt, scale=  1.00] at ( 160.20,120) {Conventional};
\node[text=drawColor,anchor=base west,inner sep=0pt, outer sep=0pt, scale=  1.00] at ( 160.20,100) {Conventional};
\node[text=drawColor,anchor=base west,inner sep=0pt, outer sep=0pt, scale=  1.00] at ( 160.20,80) {Conventional};

\node[text=drawColor,anchor=base west,inner sep=0pt, outer sep=0pt, scale=  1.00] at ( 85.20,180) {$n=100$,};
\node[text=drawColor,anchor=base west,inner sep=0pt, outer sep=0pt, scale=  1.00] at ( 85.20,160) {$n=500$,};
\node[text=drawColor,anchor=base west,inner sep=0pt, outer sep=0pt, scale=  1.00] at ( 85.20,140) {$n=1000$, };

\node[text=drawColor,anchor=base west,inner sep=0pt, outer sep=0pt, scale=  1.00] at ( 85.20, 120) {$n=100$, };
\node[text=drawColor,anchor=base west,inner sep=0pt, outer sep=0pt, scale=  1.00] at ( 85.20, 100) {$n=500$, };
\node[text=drawColor,anchor=base west,inner sep=0pt, outer sep=0pt, scale=  1.00] at ( 85.20, 80) {$n=1000$, };
\end{scope}
\end{tikzpicture}

%% file: tab3.tex
{\small
\begin{tabular}{r|rrrrrr|rrrrr}
\hline
&\multicolumn{6}{c}{No Counfounder}&\multicolumn{5}{|c}{With Confounder}\\
\cline{2-12}
&$p=2$&4&6&8&10&12&2&4&6&8&10\\
\hline
$n=100$&2&9.35&20.15&35.18&54.39&77.33&2&9.76&20.39&35.17&54.13\\
500&2&9.12&20.14&35.29&54&77&2&11.25&24.37&38.95&56.35\\
1000&2&9&20&35&54&77&2&11.33&30.98&42.48&63.44\\
\hline
\end{tabular}
}

%% file: figure008.tex
\begin{center}
\begin{tabular}{llll}
{
\setlength\unitlength{0.6mm}
\small
\begin{picture}(50,25)(0,20)
\put(0,30){\makebox(10,10){$X$}}
\put(5,35){\circle{10}}
\put(20,30){\makebox(10,10){$Y$}}
\put(25,35){\circle{10}}
\put(20,15){\framebox(10,10){$f$}}
\put(40,0){\framebox(10,10){$g$}}
\put(20,0){\makebox(10,10){$W$}}
\put(0,0){\makebox(10,10){$T$}}
\put(5,5){\circle{10}}
\put(20,5){\vector(-1,0){10}}
\put(25,5){\circle{10}}
\put(40,5){\vector(-1,0){10}}
\put(45,10){\vector(0,1){20}}
\put(42,31){\vector(-1,-2){12}}
\put(5,30){\vector(0,-1){20}}
\put(10,35){\vector(1,0){10}}
\put(40,30){\makebox(10,10){$Z$}}
\put(45,35){\circle{10}}
\put(30,35){\vector(1,0){10}}
\put(25,15){\vector(0,-11){5}}
\put(25,25){\vector(0,1){5}}
\end{picture}
}
&
{\small\begin{tabular}{l|lllll}
\multicolumn{5}{c}{LvLiNGAM}\\
\hline
&$X$
&$Y$
&$Z$
&$W$
\\
\hline
$Y$&\multicolumn{1}{l|}{$\bigcirc$}\\
\cline{3-3}
$Z$&$\bigcirc$&\multicolumn{1}{l|}{$\bigcirc$}\\
\cline{4-4}
$W$&$\bigcirc$&$\times$&\multicolumn{1}{l|}{$\times$}\\
\cline{5-5}
$T$&$\bigcirc$&$\times$&$\times$&\multicolumn{1}{l|}{$\bigcirc$}\\
\hline
\end{tabular}}&
{\small\begin{tabular}{l|lllll}
\multicolumn{5}{c}{ParceLiNGAM}\\
\hline
&$X$
&$Y$
&$Z$
&$W$
\\
\hline
$Y$&\multicolumn{1}{l|}{$\bigcirc$}\\
\cline{3-3}
$Z$&$\bigcirc$&\multicolumn{1}{l|}{$\bigcirc$}\\
\cline{4-4}
$W$&$\bigcirc$&$\times$&\multicolumn{1}{l|}{$\times$}\\
\cline{5-5}
$T$&$\bigcirc$&$\bigcirc$&$\bigcirc$&\multicolumn{1}{l|}{$\bigcirc$}\\
\hline
\end{tabular}}&
{\small\begin{tabular}{l|lllll}
\multicolumn{5}{c}{Proposed LiNGAM}\\
\hline
&$X$
&$Y$
&$Z$
&$W$
\\
\hline
$Y$&\multicolumn{1}{l|}{$\bigcirc$}\\
\cline{3-3}
$Z$&$\bigcirc$&\multicolumn{1}{l|}{$\bigcirc$}\\
\cline{4-4}
$W$&$\bigcirc$&$\bigcirc$&\multicolumn{1}{l|}{$\bigcirc$}\\
\cline{5-5}
$T$&$\bigcirc$&$\bigcirc$&$\bigcirc$&\multicolumn{1}{l|}{$\bigcirc$}\\
\hline
\end{tabular}}
\end{tabular}
\end{center}

%% file: fig_XX.tex
\begin{tikzpicture}[x=1pt,y=1pt]
\definecolor{fillColor}{RGB}{255,255,255}
\path[use as bounding box,fill=fillColor,fill opacity=0.00] (0,0) rectangle (361.35,289.08);
\begin{scope}
\path[clip] (  0.00,  0.00) rectangle (361.35,289.08);
\definecolor{drawColor}{RGB}{0,0,0}

\path[draw=drawColor,line width= 0.4pt,line join=round,line cap=round] ( 59.83, 61.20) -- (325.52, 61.20);

\path[draw=drawColor,line width= 0.4pt,line join=round,line cap=round] ( 59.83, 61.20) -- ( 59.83, 55.20);

\path[draw=drawColor,line width= 0.4pt,line join=round,line cap=round] (112.97, 61.20) -- (112.97, 55.20);

\path[draw=drawColor,line width= 0.4pt,line join=round,line cap=round] (166.11, 61.20) -- (166.11, 55.20);

\path[draw=drawColor,line width= 0.4pt,line join=round,line cap=round] (219.24, 61.20) -- (219.24, 55.20);

\path[draw=drawColor,line width= 0.4pt,line join=round,line cap=round] (272.38, 61.20) -- (272.38, 55.20);

\path[draw=drawColor,line width= 0.4pt,line join=round,line cap=round] (325.52, 61.20) -- (325.52, 55.20);

\node[text=drawColor,anchor=base,inner sep=0pt, outer sep=0pt, scale=  1.00] at ( 59.83, 39.60) {2};

\node[text=drawColor,anchor=base,inner sep=0pt, outer sep=0pt, scale=  1.00] at (112.97, 39.60) {4};

\node[text=drawColor,anchor=base,inner sep=0pt, outer sep=0pt, scale=  1.00] at (166.11, 39.60) {6};

\node[text=drawColor,anchor=base,inner sep=0pt, outer sep=0pt, scale=  1.00] at (219.24, 39.60) {8};

\node[text=drawColor,anchor=base,inner sep=0pt, outer sep=0pt, scale=  1.00] at (272.38, 39.60) {10};

\node[text=drawColor,anchor=base,inner sep=0pt, outer sep=0pt, scale=  1.00] at (325.52, 39.60) {12};

\path[draw=drawColor,line width= 0.4pt,line join=round,line cap=round] ( 49.20, 67.82) -- ( 49.20,233.26);

\path[draw=drawColor,line width= 0.4pt,line join=round,line cap=round] ( 49.20, 67.82) -- ( 43.20, 67.82);

\path[draw=drawColor,line width= 0.4pt,line join=round,line cap=round] ( 49.20,100.91) -- ( 43.20,100.91);

\path[draw=drawColor,line width= 0.4pt,line join=round,line cap=round] ( 49.20,134.00) -- ( 43.20,134.00);

\path[draw=drawColor,line width= 0.4pt,line join=round,line cap=round] ( 49.20,167.08) -- ( 43.20,167.08);

\path[draw=drawColor,line width= 0.4pt,line join=round,line cap=round] ( 49.20,200.17) -- ( 43.20,200.17);

\path[draw=drawColor,line width= 0.4pt,line join=round,line cap=round] ( 49.20,233.26) -- ( 43.20,233.26);

\node[text=drawColor,rotate= 90.00,anchor=base,inner sep=0pt, outer sep=0pt, scale=  1.00] at ( 34.80, 67.82) {0.0};

\node[text=drawColor,rotate= 90.00,anchor=base,inner sep=0pt, outer sep=0pt, scale=  1.00] at ( 34.80,100.91) {0.2};

\node[text=drawColor,rotate= 90.00,anchor=base,inner sep=0pt, outer sep=0pt, scale=  1.00] at ( 34.80,134.00) {0.4};

\node[text=drawColor,rotate= 90.00,anchor=base,inner sep=0pt, outer sep=0pt, scale=  1.00] at ( 34.80,167.08) {0.6};

\node[text=drawColor,rotate= 90.00,anchor=base,inner sep=0pt, outer sep=0pt, scale=  1.00] at ( 34.80,200.17) {0.8};

\node[text=drawColor,rotate= 90.00,anchor=base,inner sep=0pt, outer sep=0pt, scale=  1.00] at ( 34.80,233.26) {1.0};

\path[draw=drawColor,line width= 0.4pt,line join=round,line cap=round] ( 49.20, 61.20) --
	(336.15, 61.20) --
	(336.15,239.88) --
	( 49.20,239.88) --
	( 49.20, 61.20);
\end{scope}
\begin{scope}
\path[clip] (  0.00,  0.00) rectangle (361.35,289.08);
\definecolor{drawColor}{RGB}{0,0,0}

\node[text=drawColor,anchor=base,inner sep=0pt, outer sep=0pt, scale=  1.00] at (192.68, 15.60) {The Number of Variables};

\node[text=drawColor,rotate= 90.00,anchor=base,inner sep=0pt, outer sep=0pt, scale=  1.00] at ( 10.80,150.54) {Correct Rate};
\end{scope}
\begin{scope}
\path[clip] ( 49.20, 61.20) rectangle (336.15,239.88);
\definecolor{drawColor}{RGB}{223,83,107}

\path[draw=drawColor,line width= 0.4pt,line join=round,line cap=round] ( 59.83,228.30) --
	(112.97,201.83) --
	(166.11,180.32) --
	(219.24,153.85) --
	(272.38,138.96) --
	(325.52,125.72);
\definecolor{drawColor}{RGB}{97,208,79}

\path[draw=drawColor,line width= 0.4pt,line join=round,line cap=round] ( 59.83,231.61) --
	(112.97,228.30) --
	(166.11,226.64) --
	(219.24,226.64) --
	(272.38,226.64) --
	(325.52,218.37);
\definecolor{drawColor}{RGB}{34,151,230}

\path[draw=drawColor,line width= 0.4pt,line join=round,line cap=round] ( 59.83,233.26) --
	(112.97,231.61) --
	(166.11,231.61) --
	(219.24,233.26) --
	(272.38,229.95) --
	(325.52,233.26);
\definecolor{drawColor}{RGB}{223,83,107}

\path[draw=drawColor,line width= 0.4pt,dash pattern=on 4pt off 4pt ,line join=round,line cap=round] ( 59.83,224.99) --
	(112.97,191.90) --
	(166.11,186.94) --
	(219.24,152.19) --
	(272.38,130.69) --
	(325.52,100.91);
\definecolor{drawColor}{RGB}{97,208,79}

\path[draw=drawColor,line width= 0.4pt,dash pattern=on 4pt off 4pt ,line join=round,line cap=round] ( 59.83,233.26) --
	(112.97,226.64) --
	(166.11,229.95) --
	(219.24,226.64) --
	(272.38,211.75) --
	(325.52,213.41);
\definecolor{drawColor}{RGB}{34,151,230}

\path[draw=drawColor,line width= 0.4pt,dash pattern=on 4pt off 4pt ,line join=round,line cap=round] ( 59.83,233.26) --
	(112.97,233.26) --
	(166.11,231.61) --
	(219.24,233.26) --
	(272.38,228.30) --
	(325.52,228.30);
\definecolor{drawColor}{RGB}{0,0,0}

\path[draw=drawColor,line width= 0.4pt,line join=round,line cap=round] ( 49.20,145.20) rectangle (117.61, 61.20);
\definecolor{drawColor}{RGB}{223,83,107}

\path[draw=drawColor,line width= 0.4pt,line join=round,line cap=round] ( 58.20,133.20) -- ( 76.20,133.20);
\definecolor{drawColor}{RGB}{97,208,79}

\path[draw=drawColor,line width= 0.4pt,dash pattern=on 4pt off 4pt ,line join=round,line cap=round] ( 58.20,121.20) -- ( 76.20,121.20);
\definecolor{drawColor}{RGB}{34,151,230}

\path[draw=drawColor,line width= 0.4pt,line join=round,line cap=round] ( 58.20,109.20) -- ( 76.20,109.20);
\definecolor{drawColor}{RGB}{223,83,107}

\path[draw=drawColor,line width= 0.4pt,dash pattern=on 4pt off 4pt ,line join=round,line cap=round] ( 58.20, 97.20) -- ( 76.20, 97.20);
\definecolor{drawColor}{RGB}{97,208,79}

\path[draw=drawColor,line width= 0.4pt,line join=round,line cap=round] ( 58.20, 85.20) -- ( 76.20, 85.20);
\definecolor{drawColor}{RGB}{34,151,230}

\path[draw=drawColor,line width= 0.4pt,dash pattern=on 4pt off 4pt ,line join=round,line cap=round] ( 58.20, 73.20) -- ( 76.20, 73.20);
\definecolor{drawColor}{RGB}{0,0,0}

\node[text=drawColor,anchor=base west,inner sep=0pt, outer sep=0pt, scale=  1.00] at ( 85.20,129.76) {100,J};

\node[text=drawColor,anchor=base west,inner sep=0pt, outer sep=0pt, scale=  1.00] at ( 85.20,117.76) {500,J};

\node[text=drawColor,anchor=base west,inner sep=0pt, outer sep=0pt, scale=  1.00] at ( 85.20,105.76) {1000,J};

\node[text=drawColor,anchor=base west,inner sep=0pt, outer sep=0pt, scale=  1.00] at ( 85.20, 93.76) {100,I};

\node[text=drawColor,anchor=base west,inner sep=0pt, outer sep=0pt, scale=  1.00] at ( 85.20, 81.76) {500,I};

\node[text=drawColor,anchor=base west,inner sep=0pt, outer sep=0pt, scale=  1.00] at ( 85.20, 69.76) {1000,I};
\end{scope}
\begin{scope}
\path[clip] (  0.00,  0.00) rectangle (361.35,289.08);
\definecolor{drawColor}{RGB}{0,0,0}

\node[text=drawColor,anchor=base,inner sep=0pt, outer sep=0pt, scale=  1.20] at (192.68,260.34) {\bfseries Precision};
\end{scope}
\end{tikzpicture}

%% file: fig_YY.tex
\begin{tikzpicture}[x=1pt,y=1pt]
\definecolor{fillColor}{RGB}{255,255,255}
\path[use as bounding box,fill=fillColor,fill opacity=0.00] (0,0) rectangle (361.35,289.08);
\begin{scope}
\path[clip] (  0.00,  0.00) rectangle (361.35,289.08);
\definecolor{drawColor}{RGB}{0,0,0}

\path[draw=drawColor,line width= 0.4pt,line join=round,line cap=round] ( 59.83, 61.20) -- (325.52, 61.20);

\path[draw=drawColor,line width= 0.4pt,line join=round,line cap=round] ( 59.83, 61.20) -- ( 59.83, 55.20);

\path[draw=drawColor,line width= 0.4pt,line join=round,line cap=round] (126.25, 61.20) -- (126.25, 55.20);

\path[draw=drawColor,line width= 0.4pt,line join=round,line cap=round] (192.68, 61.20) -- (192.68, 55.20);

\path[draw=drawColor,line width= 0.4pt,line join=round,line cap=round] (259.10, 61.20) -- (259.10, 55.20);

\path[draw=drawColor,line width= 0.4pt,line join=round,line cap=round] (325.52, 61.20) -- (325.52, 55.20);

\node[text=drawColor,anchor=base,inner sep=0pt, outer sep=0pt, scale=  1.00] at ( 59.83, 39.60) {2};

\node[text=drawColor,anchor=base,inner sep=0pt, outer sep=0pt, scale=  1.00] at (126.25, 39.60) {4};

\node[text=drawColor,anchor=base,inner sep=0pt, outer sep=0pt, scale=  1.00] at (192.68, 39.60) {6};

\node[text=drawColor,anchor=base,inner sep=0pt, outer sep=0pt, scale=  1.00] at (259.10, 39.60) {8};

\node[text=drawColor,anchor=base,inner sep=0pt, outer sep=0pt, scale=  1.00] at (325.52, 39.60) {10};

\path[draw=drawColor,line width= 0.4pt,line join=round,line cap=round] ( 49.20, 67.82) -- ( 49.20,233.26);

\path[draw=drawColor,line width= 0.4pt,line join=round,line cap=round] ( 49.20, 67.82) -- ( 43.20, 67.82);

\path[draw=drawColor,line width= 0.4pt,line join=round,line cap=round] ( 49.20,109.18) -- ( 43.20,109.18);

\path[draw=drawColor,line width= 0.4pt,line join=round,line cap=round] ( 49.20,150.54) -- ( 43.20,150.54);

\path[draw=drawColor,line width= 0.4pt,line join=round,line cap=round] ( 49.20,191.90) -- ( 43.20,191.90);

\path[draw=drawColor,line width= 0.4pt,line join=round,line cap=round] ( 49.20,233.26) -- ( 43.20,233.26);

\node[text=drawColor,rotate= 90.00,anchor=base,inner sep=0pt, outer sep=0pt, scale=  1.00] at ( 34.80, 67.82) {0};

\node[text=drawColor,rotate= 90.00,anchor=base,inner sep=0pt, outer sep=0pt, scale=  1.00] at ( 34.80,109.18) {200};

\node[text=drawColor,rotate= 90.00,anchor=base,inner sep=0pt, outer sep=0pt, scale=  1.00] at ( 34.80,150.54) {400};

\node[text=drawColor,rotate= 90.00,anchor=base,inner sep=0pt, outer sep=0pt, scale=  1.00] at ( 34.80,191.90) {600};

\node[text=drawColor,rotate= 90.00,anchor=base,inner sep=0pt, outer sep=0pt, scale=  1.00] at ( 34.80,233.26) {800};

\path[draw=drawColor,line width= 0.4pt,line join=round,line cap=round] ( 49.20, 61.20) --
	(336.15, 61.20) --
	(336.15,239.88) --
	( 49.20,239.88) --
	( 49.20, 61.20);
\end{scope}
\begin{scope}
\path[clip] (  0.00,  0.00) rectangle (361.35,289.08);
\definecolor{drawColor}{RGB}{0,0,0}

\node[text=drawColor,anchor=base,inner sep=0pt, outer sep=0pt, scale=  1.00] at (192.68, 15.60) {The Number of Variables};

\node[text=drawColor,rotate= 90.00,anchor=base,inner sep=0pt, outer sep=0pt, scale=  1.00] at ( 10.80,150.54) {The Number of Calculations};
\end{scope}
\begin{scope}
\path[clip] ( 49.20, 61.20) rectangle (336.15,239.88);
\definecolor{drawColor}{RGB}{223,83,107}

\path[draw=drawColor,line width= 0.4pt,line join=round,line cap=round] ( 59.83, 68.23) --
	(126.25, 69.87) --
	(192.68, 72.04) --
	(259.10, 75.09) --
	(325.52, 79.09);
\definecolor{drawColor}{RGB}{97,208,79}

\path[draw=drawColor,line width= 0.4pt,line join=round,line cap=round] ( 59.83, 68.23) --
	(126.25, 69.93) --
	(192.68, 72.39) --
	(259.10, 75.64) --
	(325.52, 79.51);
\definecolor{drawColor}{RGB}{34,151,230}

\path[draw=drawColor,line width= 0.4pt,line join=round,line cap=round] ( 59.83, 68.23) --
	(126.25, 69.90) --
	(192.68, 73.30) --
	(259.10, 76.09) --
	(325.52, 80.11);
\definecolor{drawColor}{RGB}{223,83,107}

\path[draw=drawColor,line width= 0.4pt,dash pattern=on 4pt off 4pt ,line join=round,line cap=round] ( 59.83, 68.23) --
	(126.25, 70.42) --
	(192.68, 80.30) --
	(259.10,131.77) --
	(294.34,289.08);
\definecolor{drawColor}{RGB}{97,208,79}

\path[draw=drawColor,line width= 0.4pt,dash pattern=on 4pt off 4pt ,line join=round,line cap=round] ( 59.83, 68.23) --
	(126.25, 70.22) --
	(192.68, 76.66) --
	(259.10,100.32) --
	(325.52,216.05);
\definecolor{drawColor}{RGB}{34,151,230}

\path[draw=drawColor,line width= 0.4pt,dash pattern=on 4pt off 4pt ,line join=round,line cap=round] ( 59.83, 68.23) --
	(126.25, 70.12) --
	(192.68, 76.13) --
	(259.10, 92.02) --
	(325.52,169.51);
\definecolor{drawColor}{RGB}{0,0,0}

\path[draw=drawColor,line width= 0.4pt,line join=round,line cap=round] ( 49.20,239.88) rectangle (117.61,155.88);
\definecolor{drawColor}{RGB}{223,83,107}

\path[draw=drawColor,line width= 0.4pt,line join=round,line cap=round] ( 58.20,227.88) -- ( 76.20,227.88);
\definecolor{drawColor}{RGB}{97,208,79}

\path[draw=drawColor,line width= 0.4pt,dash pattern=on 4pt off 4pt ,line join=round,line cap=round] ( 58.20,215.88) -- ( 76.20,215.88);
\definecolor{drawColor}{RGB}{34,151,230}

\path[draw=drawColor,line width= 0.4pt,line join=round,line cap=round] ( 58.20,203.88) -- ( 76.20,203.88);
\definecolor{drawColor}{RGB}{223,83,107}

\path[draw=drawColor,line width= 0.4pt,dash pattern=on 4pt off 4pt ,line join=round,line cap=round] ( 58.20,191.88) -- ( 76.20,191.88);
\definecolor{drawColor}{RGB}{97,208,79}

\path[draw=drawColor,line width= 0.4pt,line join=round,line cap=round] ( 58.20,179.88) -- ( 76.20,179.88);
\definecolor{drawColor}{RGB}{34,151,230}

\path[draw=drawColor,line width= 0.4pt,dash pattern=on 4pt off 4pt ,line join=round,line cap=round] ( 58.20,167.88) -- ( 76.20,167.88);
\definecolor{drawColor}{RGB}{0,0,0}

\node[text=drawColor,anchor=base west,inner sep=0pt, outer sep=0pt, scale=  1.00] at ( 85.20,224.44) {100,J};

\node[text=drawColor,anchor=base west,inner sep=0pt, outer sep=0pt, scale=  1.00] at ( 85.20,212.44) {500,J};

\node[text=drawColor,anchor=base west,inner sep=0pt, outer sep=0pt, scale=  1.00] at ( 85.20,200.44) {1000,J};

\node[text=drawColor,anchor=base west,inner sep=0pt, outer sep=0pt, scale=  1.00] at ( 85.20,188.44) {100,I};

\node[text=drawColor,anchor=base west,inner sep=0pt, outer sep=0pt, scale=  1.00] at ( 85.20,176.44) {500,I};

\node[text=drawColor,anchor=base west,inner sep=0pt, outer sep=0pt, scale=  1.00] at ( 85.20,164.44) {1000,I};
\end{scope}
\begin{scope}
\path[clip] (  0.00,  0.00) rectangle (361.35,289.08);
\definecolor{drawColor}{RGB}{0,0,0}

\node[text=drawColor,anchor=base,inner sep=0pt, outer sep=0pt, scale=  1.20] at (192.68,260.34) {\bfseries Efficiency};
\end{scope}
\end{tikzpicture}

%% file: figure090.tex
\setlength\unitlength{0.40mm}
\footnotesize
\begin{center}
\begin{picture}(195,130)(0,00)

\put(15,15){\oval(30,28)}
\put(5,15){\makebox(20,10){Positive}}
\put(5,5){\makebox(20,10){X-ray}}

\put(60,40){\oval(30,28)}
\put(50,40){\makebox(20,10){Either}}
\put(50,30){\makebox(20,10){T or L}}

\put(95,75){\oval(30,28)}
\put(85,75){\makebox(20,10){Lung}}
\put(85,65){\makebox(20,10){Cancer}}

\put(135,110){\oval(35,25)}
\put(125,105){\makebox(20,10){Smoking}}

\put(125,30){\oval(35,25)}
\put(115,25){\makebox(20,10){Dyspnoac}}

\put(175,75){\oval(35,25)}
\put(165,70){\makebox(20,10){Bronchitis}}

\put(15,70){\oval(40,20)}
\put(5,65){\makebox(20,10){Tuberculosis}}

\put(15,115){\oval(30,28)}
\put(5,115){\makebox(20,10){Visit}}
\put(5,105){\makebox(20,10){Asia}}

\thicklines
\put(14,28){\color{red}\vector(4,3){64}}

\put(15,100){\vector(0,-1){18}}
\put(12,100){\color{red}\vector(0,-1){18}}
\put(15,60){\color{red}\vector(0,-1){30}}
\put(29,61){\vector(1,-1){17}}
\put(45,38){\vector(-1,-1){17}}
\put(82,67){\vector(-1,-1){15}}
\put(84,63){\color{red}\vector(-1,-1){14}}

\put(118,103){\vector(-1,-1){15}}

\put(146,100){\vector(1,-1){16}}

\put(165,87){\color{red}\vector(-1,1){17}}

\put(165,65){\vector(-1,-1){25}}
\put(143,38){\color{red}\vector(1,1){25}}

\put(75,40){\vector(3,-1){30}}
\put(75,37){\color{red}\vector(3,-1){31}}

\end{picture}
\end{center}